\documentclass[11pt]{cambrian}
\usepackage[utf8]{inputenc}
\usepackage[T1]{fontenc}
\usepackage{hyperref}
\usepackage{url}
\usepackage{booktabs}
\usepackage{amsfonts}
\usepackage{amsmath}
\usepackage{amssymb}
\usepackage{amsthm}
\usepackage{nicefrac}
\usepackage{microtype}
\usepackage{xcolor}
\usepackage{graphicx}
\usepackage{subcaption}
\usepackage{float}
\usepackage{multirow}
\usepackage{makecell}
\usepackage{enumitem}
\usepackage{algorithm}
\usepackage{algorithmic}
\usepackage{tikz}
\usetikzlibrary{positioning,arrows.meta,calc,fit,backgrounds}
\usepackage[capitalize,noabbrev,nameinlink]{cleveref}

\usepackage{pifont}
\usepackage{xspace}

\newcommand{\ourmethod}{Sensorimotor Alignment\xspace}
\newcommand{\PA}{$\mathrm{S}^2$\xspace}

\newcommand{\customsize}{\fontsize{9pt}{11pt}\selectfont}

\usepackage{caption}
\DeclareCaptionFont{custom}{\customsize}

\newif\ifshownotes
\shownotesfalse  %
\usepackage{todonotes}

\usepackage[most]{tcolorbox}
\usepackage{marginnote}
\usepackage{placeins}
\usepackage{needspace}
\usepackage{etoolbox}
\usepackage{soul}
\usepackage[numbers,sort&compress]{natbib}
\definecolor{bluelink}{RGB}{0,113,188}
\hypersetup{colorlinks=true, citecolor=green!93!black, linkcolor=black, urlcolor=bluelink}
\newcommand{\github}{\raisebox{-1.5pt}{\includegraphics[height=1.05em]{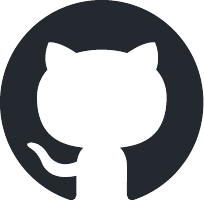}}\xspace}
\newcommand{\worldwideweb}{\raisebox{-1.5pt}{\includegraphics[height=1.05em]{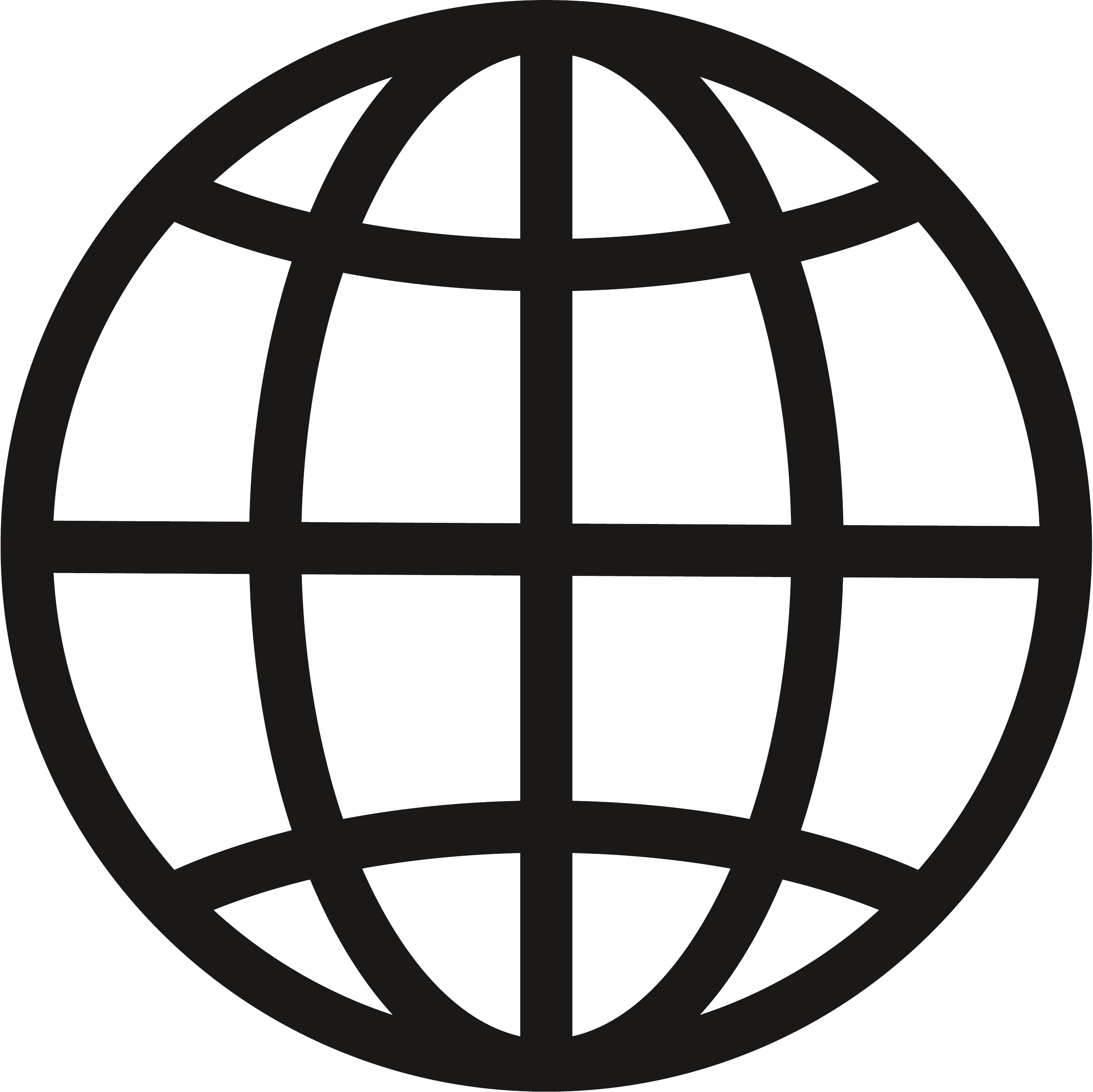}}\xspace}

\newcommand{\SO}{\mathrm{SO}}

\title{Sensorimotor Alignment: \\
A Platonic Proxy for Embodied Policy Success}

\author{%
Yunfei~Xie\textsuperscript{1},
Howard~H.~Qian\textsuperscript{1},
Shiyi~Lan\textsuperscript{2},
Kaiyu~Hang\textsuperscript{1},
Chen~Wei\textsuperscript{1\textdagger}\\
\textsuperscript{1}Rice University,\quad \textsuperscript{2}NVIDIA
}

\begin{abstract}

Robot foundation models are typically evaluated through costly closed-loop rollouts, motivating low-cost offline proxies for model capability.
The Platonic Representation Hypothesis (PRH)~\citep{huh2024platonic} suggests that as representations improve, modalities such as vision and language converge toward a shared statistical model of reality. We ask whether motion trajectories form another such modality: as embodied models become more capable, should their sensory representations converge toward the structure of physical motion?
To test this, we propose \textbf{Sensorimotor Alignment} (\PA), a training-free offline metric that measures how well a model's sensory representation aligns with the geometry of physical motion.
We find that \PA correlates positively with downstream robot task success across model families, making it a low-cost offline proxy for prioritizing models before closed-loop evaluation. \PA can also identify promising vision-language model (VLM) backbones \textit{before} costly fine-tuning into vision-language-action models (VLAs).
Beyond this correlation, \PA offers representation-level explanations for phenomena observed in VLAs and world action models (WAMs).
With \PA, we extend PRH to robotics and make sensorimotor alignment measurable, offering a low-cost, representation-level view of robot foundation models that closed-loop success rates alone cannot reveal.

\end{abstract}

\begin{document}

\maketitle
{\renewcommand{\thefootnote}{}\footnotetext{\textsuperscript{\textdagger}Corresponding author.}}

\begin{center}
    \renewcommand{\arraystretch}{1.2}
    \begin{tabular}{rll}
        \worldwideweb & \textbf{Website} & \url{https://yunfeixie233.github.io/s2/}\\
        \github & \textbf{Code} & \url{https://github.com/yunfeixie233/sensorimotor-alignment}\\
    \end{tabular}
\end{center}

\clearpage
\begin{figure}[!htbp]
  \centering

  \includegraphics[width=\linewidth]{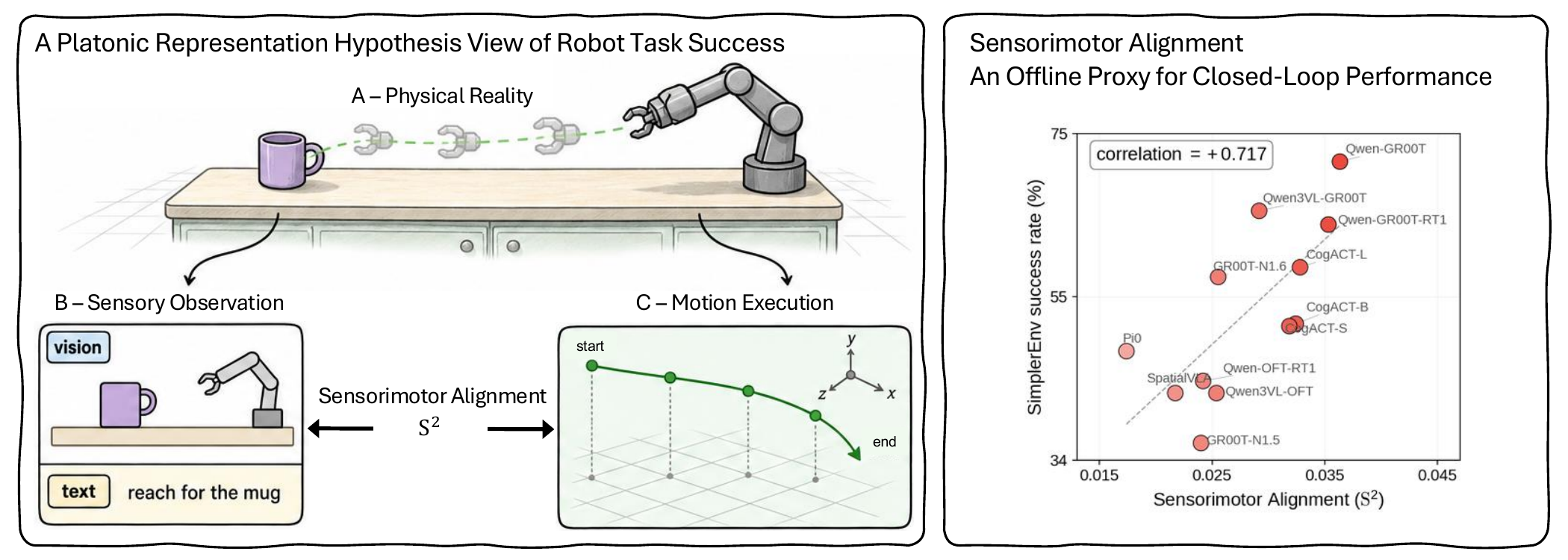}

  \caption[Sensorimotor Alignment tests a PRH-inspired view of embodiment.]{\textbf{Left:} We test the Platonic Representation Hypothesis (PRH) in robotics, using image--text features as the sensory representation and end-effector trajectories as the motor representation: a more capable embodied model should have sensory representations that align more closely with this motor representation. To validate this, we define \textbf{\ourmethod} (\PA).
\textbf{Right:} We measure \PA across 12 vision-language-action models (VLAs) and find that \PA correlates positively with SimplerEnv task success rate (Pearson $r=+0.717$), providing empirical support for this PRH-inspired view of embodiment.\label{fig:teaser}}

\end{figure}

\section{Introduction}
\label{sec:intro}

Robot foundation models increasingly build on large pretrained vision-language models~\citep{beyer2024paligemma,bai2025qwen25vltechnicalreport}, video models~\citep{wan21}, and world action models~\citep{feng2025vidar,li2026causal,bi2025motus}. Yet selecting and evaluating these models remains expensive: the most reliable test is still closed-loop rollout in simulation~\citep{li2024simplerenv,mu2025robotwin} or the real world~\citep{atreya2025roboarena,chen2026manipulationnet}, where evaluation is slow, noisy, and benchmark-dependent. This expense motivates low-cost offline proxies for policy capability.

A natural approach is to use existing perception-side proxies, such as visual question answering scores for VLMs~\citep{liu2023mmbench,yue2024mmmu} or pixel-level video quality for world models~\citep{heusel2017gans,wang2004image,jayasumana2024rethinking}. However, these metrics measure passive perception or generation quality, not whether a representation is organized for action~\citep{vlm4vla,kachaev2025don,tiancontrol,pai2025mimic}. Two scenes may look visually different but require similar robot motions; conversely, two visually similar scenes may demand different trajectories. For robot task success, a model must produce different motions across tasks even in similar scenes. This requires aligning the sensory representation with the motor representation.

This perspective connects robotics to the Platonic Representation Hypothesis (PRH)~\citep{huh2024platonic}, which argues that as models of different modalities scale and solve diverse tasks, they are converging to a shared statistical model of reality in each of their own representation spaces.
For example, images and text are two projections of the same underlying world.
As vision and language models scale,
they increasingly agree on the relative distances between data points.
We bring PRH to robotics and ask: can physical robot motion be viewed as another projection of that same world, such that more capable embodied models better align their vision and language representations with the structure of motion trajectories?

We test this prediction of PRH: the more capable an embodied model is, the more closely its sensory representations align with its motor representation. In other words, episodes nearby in sensory representation space should also be nearby in motor representation space.

We formalize this as \textbf{Sensorimotor Alignment} (\PA), a training-free metric, illustrated in \cref{fig:teaser}. For each episode in a robot dataset, a model's image and text features form its sensory representation, and the recorded end-effector trajectory forms its motor representation. We measure alignment by comparing each episode's nearest neighbors in sensory and motor spaces. More overlap between the two neighbor sets means stronger sensorimotor alignment.

In \cref{sec:pa-predicts-sr}, we show that \PA correlates positively with downstream robot task success across model families, in simulated and real-world settings. The correlation is strong for Vision-Language-Action models (VLAs) in closed-loop manipulation, indicating that models with higher sensorimotor alignment tend to perform better. \PA computed on raw Vision-Language Models (VLMs) also correlates with the task success of the resulting VLAs, more strongly than their visual question answering (VQA) scores do. For World Action Models (WAMs), \PA likewise correlates more strongly with task success than video-quality metrics such as FID~\citep{heusel2017gans}, SSIM~\citep{wang2004image}, and PSNR~\citep{hore2010image}, and for real-robot policies, \PA computed offline correlates with physical-robot task success on a shared set of tasks. Across these settings, sensorimotor alignment correlates with task success more closely than VQA or video-quality metrics.

Prior work has observed several phenomena in VLAs and WAMs only through closed-loop results, without explaining them at the representation level. We use \mbox{\PA} to supply that explanation, as shown in \cref{sec:pa-reveals}. For example, fine-tuned VLAs often ignore the language instruction and act mainly on what they see; we find that their task success correlates with the \mbox{\PA} of their vision features, not their text features.

Our contributions are threefold:
\begin{itemize}
[leftmargin=*,itemsep=0mm]

    \item Inspired by the Platonic Representation Hypothesis, we propose a new view of embodiment: more capable embodied models should align their sensory representations more closely with the structure of motion trajectories. To test this, we introduce \PA, a training-free offline metric that measures alignment between sensory and motor representations.
    \item Across model families, in simulation and on real robots, \PA correlates positively with task success, making it a low-cost offline proxy for choosing models before closed-loop evaluation.
    \item Furthermore, we use \PA to explain, at the representation level, several phenomena in VLAs and WAMs that were previously observed only through closed-loop results, making it a diagnostic tool for understanding how design choices shape embodied models.

\end{itemize}

We make sensorimotor alignment measurable, offering a low-cost, representation-level view of robot foundation models that closed-loop success rates alone cannot reveal.

\section{Measuring \ourmethod}
\label{sec:pa}

\label{subsec:pa-intuition}

Motivated by the Platonic Representation Hypothesis (PRH)~\citep{huh2024platonic}, we hypothesize that the more capable an embodied model is, the more closely its sensory representations align with its motor representation. Here, sensory representations are the model's image and text features, and each episode's end-effector trajectory is its motor representation.

To make this alignment measurable, we define \textbf{Sensorimotor Alignment} (\PA) by constructing both representations and comparing their neighborhoods, as follows:

\begin{figure}[!t]

  \centering
  \includegraphics[width=\textwidth]{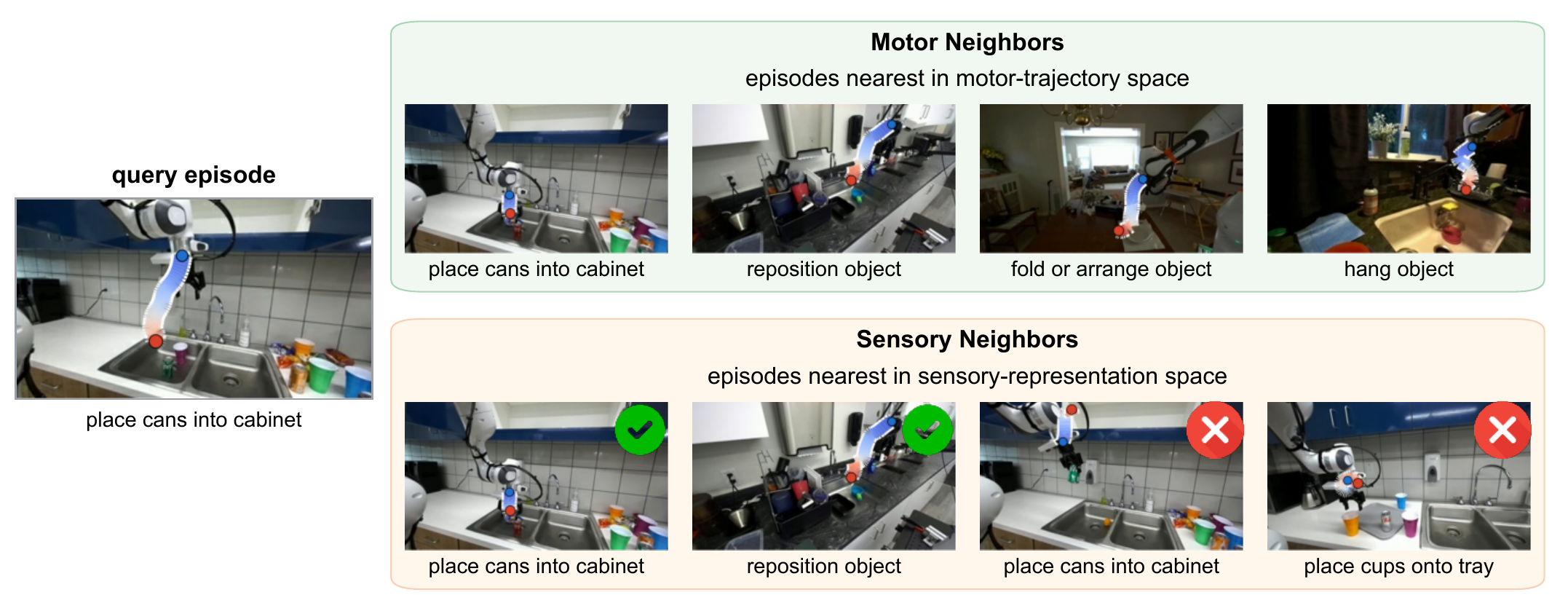}
\caption{\textbf{Measuring Sensorimotor Alignment (\PA) by nearest-neighbor overlap.}
For the query episode (left), sensory neighbors (bottom right) have similar vision-text features, and motor neighbors (top right) have similar trajectories.
{\color[RGB]{0,177,38}Green checks} mark shared neighbors; {\color[RGB]{242,59,53}red crosses} mark neighbors found only in the sensory set.
A larger overlap means that episodes with similar sensory representations also follow similar motor trajectories.}

  \label{fig:pa-worked-example}
\end{figure}

\textbf{Reference dataset.}
To measure cross-modal alignment, we use the DROID~\citep{khazatsky2024droid} and BridgeDataV2~\citep{walke2023bridgedata} datasets, which pair sensory inputs with physical motions. From a dataset, we sample $N$ episodes and compute both representations from each episode. For episode $i$, we randomly sample one image frame and pair it with the task instruction; this image--text pair forms the sensory input $\mathbf{x}_i$. Sampling details are provided in Appendix~\ref{app:fixed-samples}.

\begin{samepage}
\textbf{Sensory representation.}
For each episode $i$, we encode the image frame and text instruction with the vision and text encoders of a robot foundation model, obtaining a sensory embedding
\[
\mathbf{z}_i=f(\mathbf{x}_i)\in\mathbb{R}^{d},
\]
where $f$ applies the vision encoder to the image and the text encoder to the instruction, and we average-pool the output token features into a single $d$-dimensional vector.\par
\end{samepage}

\textbf{Motor representation.}
For the sampled frame in $\mathbf{x}_i$, we extract end-effector positions, orientations, and gripper states over the next $h$ steps, forming episode $i$'s motor representation:
\[
\boldsymbol{\tau}_i=(x_i,y_i,z_i,\phi_i,\theta_i,\psi_i,g_i)\in\mathbb{R}^{h\times7},
\]
where row $u$ is the state $u$ steps after the sampled frame: $x,y,z$ are position coordinates, $\phi,\theta,\psi$ are roll, pitch, and yaw angles, and $g$ is the recorded gripper coordinate. We encode positions as vectors in $\mathbb{R}^3$ and orientation angles as rotations in $\SO(3)$ so that distances reflect the geometry of translations and rotations. Details are in Appendix~\ref{app:base-frame}.

\textbf{Sensorimotor alignment.}
We measure alignment with a mutual nearest-neighbor metric comparing the $k$-nearest-neighbor sets in sensory and motor spaces; greater overlap indicates stronger alignment between an episode's sensory and motor representations.

We use cosine distance in sensory space and Dynamic Time Warping (DTW)~\citep{sakoe1978dynamic,qian2026robotok} in motor space to compare trajectories at different speeds. For episodes $i$ and $j$:
\[
d_{\mathrm{sensory}}(i,j)=1-\cos(\mathbf{z}_i,\mathbf{z}_j),
\qquad
d_{\mathrm{motor}}(i,j)=\operatorname{DTW}(\boldsymbol{\tau}_i,\boldsymbol{\tau}_j).
\]
Smaller distances mean closer neighbors. Appendices~\ref{app:base-frame}--\ref{app:cknna-estimator} detail DTW and normalization.

\begin{samepage}
For each episode $i$, we use $d_{\mathrm{sensory}}$ to form the sensory $k$-nearest-neighbor set $\mathcal{N}_{\mathrm{sensory}}(i)$ and $d_{\mathrm{motor}}$ to form the motor $k$-nearest-neighbor set $\mathcal{N}_{\mathrm{motor}}(i)$. We then take the intersection of the two sets, divide its size by $k$, and average over all $N$ episodes~\citep{huh2024platonic}:
\begin{equation}
m_{\mathrm{NN}}=\frac{1}{Nk}\sum_{i=1}^{N}\left|\mathcal{N}_{\mathrm{sensory}}(i)\cap\mathcal{N}_{\mathrm{motor}}(i)\right|.
\label{eq:neighbor-overlap}
\end{equation}
Greater overlap indicates that episodes with similar sensory representations also have similar motor trajectories, as shown in \cref{fig:pa-worked-example}.\par
\end{samepage}

In practice, our reported \PA uses a centered, normalized variant of CKNNA with $k=10$ neighbors unless stated otherwise; details are given in Appendix~\ref{app:cknna-estimator}.

\FloatBarrier
\section{\ourmethod\ \texorpdfstring{Correlates with}{Correlates with} Robot Task Success Rate}
\label{sec:pa-predicts-sr}

The Platonic Representation Hypothesis (PRH)~\citep{huh2024platonic} argues that as models of different modalities become more capable, their representations converge toward a shared statistical model of reality. While the evidence for PRH comes from observational data such as images and text, robotics adds the motion the robot executes, which an embodied model must produce rather than observe. We test the prediction that models whose image--text representations align more closely with end-effector trajectories should achieve higher downstream task success.

We quantify this alignment using \ourmethod (\PA), which compares episode neighborhoods in sensory and motor space, as defined in \cref{sec:pa}. A positive correlation between \PA and task success would provide empirical support for this PRH-inspired view of embodiment. It would also support using \PA as a low-cost offline proxy for deciding which models to evaluate in closed loop first.

\cref{sec:finding-pa-predicts-across-families} establishes this positive correlation across 3 model families and real-robot policies. \cref{sec:alternative-metrics} shows that \PA correlates with task success more closely than VQA and video-quality metrics. \cref{sec:finding-robustness} shows that the correlation holds across computation settings and drops in a negative control with a mismatched embodiment.

\subsection{\PA Correlates with Task Success Across Model Families}\label{sec:finding-pa-predicts-across-families}

In \cref{fig:pa-predicts-sr-1x3}, we test 3 model families and real-robot policies, comparing each model's \mbox{\PA} on a reference dataset (\mbox{$x$}-axis) with its downstream manipulation success rate (\mbox{$y$}-axis).

\begingroup\setlength{\intextsep}{2pt}
\begin{figure}[!htbp]

  \centering
  \begin{tikzpicture}
    \node[inner sep=0,outer sep=0,anchor=south west] (img) {\includegraphics[width=\linewidth]{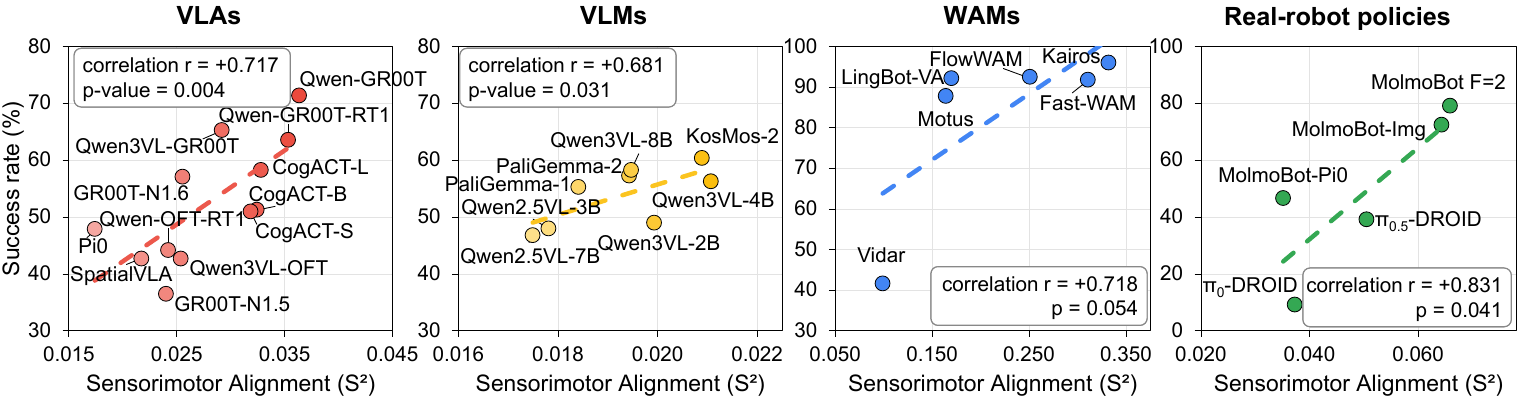}};
    \begin{scope}[font=\fontsize{7}{8}\selectfont\bfseries]
      \node[anchor=base east,inner sep=0,outer sep=0] at ([xshift=0.10982667\linewidth,yshift=0.24551157\linewidth]img.south west) {(a)};
      \node[anchor=base east,inner sep=0,outer sep=0] at ([xshift=0.37497752\linewidth,yshift=0.24551157\linewidth]img.south west) {(b)};
      \node[anchor=base east,inner sep=0,outer sep=0] at ([xshift=0.61376114\linewidth,yshift=0.24551157\linewidth]img.south west) {(c)};
      \node[anchor=base east,inner sep=0,outer sep=0] at ([xshift=0.79808534\linewidth,yshift=0.24477230\linewidth]img.south west) {(d)};
    \end{scope}
  \end{tikzpicture}
  \par
  \caption{\textbf{Sensorimotor Alignment (\PA) correlates with task success across model families and in simulated and real-world evaluations.}
Each panel plots offline \PA ($x$-axis) against downstream task success rate ($y$-axis) for (a) fine-tuned vision-language-action models (VLAs), (b) raw vision-language models (VLMs), (c) world action models (WAMs), and (d) real-robot policies.
For raw VLMs, task success is measured after VLA fine-tuning.
These correlations (Pearson's $r$; unadjusted one-tailed $p$-values) support \PA as an offline proxy across model families.\label{fig:pa-predicts-sr-1x3}}

\end{figure}\endgroup

\textbf{Vision-language-action models (VLAs).}\label{sec:finding-vla-sr}
We test 12 VLAs against their SimplerEnv~\citep{li2024simplerenv} success rates. In \mbox{\cref{fig:pa-predicts-sr-1x3}a}, the Pearson correlation \mbox{$r\!=\!{+}0.717$} is strong, with a model-bootstrap \mbox{$95\%$} confidence interval of \mbox{$[{+}0.436,\,{+}0.912]$}.
Thus, models with higher sensorimotor alignment tend to achieve higher task success rates. The correlation holds across reference datasets, model families, scales, and VLA training setups, as shown in \cref{fig:cross-embodiment-vla} and Appendix~\ref{app:confound-controls}.

\textbf{Vision-language models (VLMs).}\label{sec:finding-vlm-predict-vla}
We ask whether \mbox{\PA} computed on a raw VLM, before any robot fine-tuning, correlates with the task success of the VLA that is later fine-tuned from that backbone. We compute \mbox{\PA} for 8 raw VLM backbones from VLM4VLA~\citep{vlm4vla} and compare it with the SimplerEnv~\citep{li2024simplerenv} success rates of the resulting VLAs. We find a strong correlation, \mbox{$r=+0.681$} with one-tailed \mbox{$p=0.031$}, as shown in \mbox{\cref{fig:pa-predicts-sr-1x3}b}. \mbox{\PA} can therefore identify promising VLM backbones before costly VLA fine-tuning. Appendix~\ref{app:models} lists the backbones and \mbox{\PA} settings.

\textbf{World action models (WAMs).}\label{sec:finding-va-sr}
\leavevmode We also test WAMs, which build on video generators rather than VLMs. We compute \mbox{\PA} for 6 WAMs on the bimanual ALOHA dataset~\citep{zhao2023learning} and compare it with their success rates on RoboTwin~2.0~\citep{mu2025robotwin}. For WAMs, the sensory representation is the video-generator latent used for action decoding, and all 6 models share the same ALOHA trajectories and \mbox{\PA} settings. \mbox{\PA} again correlates strongly, \mbox{$r=+0.718$} with one-tailed \mbox{$p=0.054$}, as shown in \mbox{\cref{fig:pa-predicts-sr-1x3}c}. Appendix~\ref{app:wam-protocol} lists the 6~WAMs and shared evaluation protocol.

\textbf{Real-robot policies.}
We test whether the correlation observed in simulation extends to real robots. Because real-world success rates depend on many factors, such as the task set, robot, cameras, and protocol, we take them all from one shared evaluation: the physical-robot results that MolmoB0T~\citep{deshpande2026molmobot} reports for 5 policies on 40 tasks. We then compute \PA for their public checkpoints on the same DROID reference dataset. We find a strong correlation with real-robot success, \mbox{$r=+0.831$} with one-tailed \mbox{$p=0.041$}, as shown in \mbox{\cref{fig:pa-predicts-sr-1x3}d}. Appendix~\ref{app:real-robot-protocol} describes the shared evaluation conditions and checkpoint scoring.

We further validate that \mbox{\PA} correlates with downstream task success for frozen vision backbones under low-data behavior cloning and visual-reward learning, as shown in \mbox{Appendix~\ref{app:hu-backbones}}.

\FloatBarrier

\subsection{\PA Correlates with Task Success More Closely Than Alternative Metrics}
\label{sec:alternative-metrics}

\begin{sloppypar}
Prior work asks whether offline scores correlate with downstream
task performance.
VLM4VLA~\citep{vlm4vla} examines the correlation between VQA
benchmark scores and the performance of the resulting VLA policies,
finding weak or even negative correlations.
For video models, \citet{tiancontrol} find that the quality of predicted video agrees inconsistently with downstream performance across tasks.
A common limitation is that these metrics measure language understanding
or pixel-level video quality, neither of which depends on the motion the robot must execute.
\PA instead compares the nearest-neighbor sets of sensory embeddings and motor trajectories.
As we show below, \PA correlates more strongly with task success rate
than any VQA or video-quality metric we evaluate.
\end{sloppypar}

\begin{figure}[!htbp]

  \captionsetup{justification=raggedright}
  \centering
  \begin{minipage}{0.66\textwidth}
    \includegraphics[width=\linewidth]{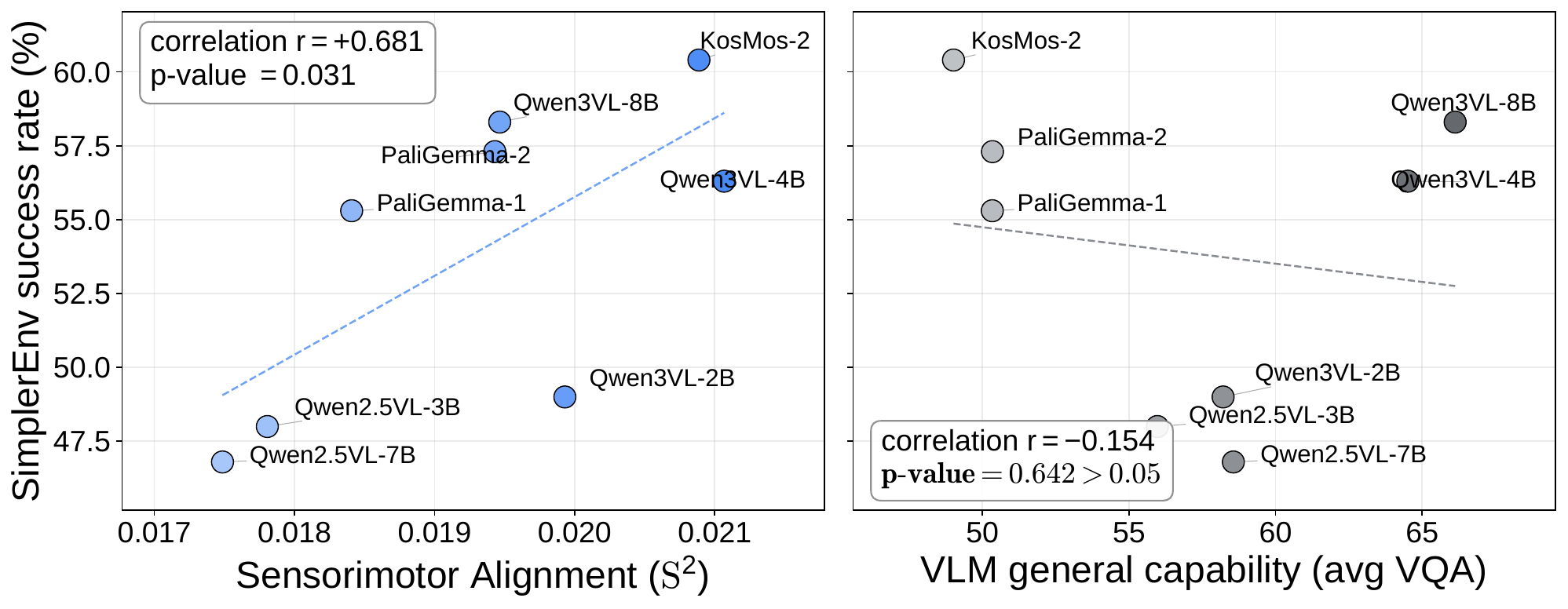}
  \end{minipage}\hfill
  \begin{minipage}{0.3\textwidth}
    \caption{\textbf{Sensorimotor Alignment (\PA) correlates with VLA success rate where VQA scores do not.}
Across 8 raw vision-language models (VLMs)~\citep{vlm4vla}, \PA (left) correlates strongly with SimplerEnv task success after fine-tuning into vision-language-action models (VLAs), whereas visual question answering (VQA) scores (right) show virtually no correlation.\label{fig:pa-vs-vqa-simpler}}
  \end{minipage}

\end{figure}

\textbf{\PA correlates with VLA success rate where VQA scores do not.}
We compare \mbox{\PA} with aggregate VQA scores as offline proxies, using the same 8 raw VLM backbones and evaluation setting as in \cref{sec:finding-vlm-predict-vla}.
We find that VQA scores show virtually no correlation with VLA success rate, at $r=-0.154$ (right panel of \cref{fig:pa-vs-vqa-simpler}). \mbox{\PA}, by contrast, correlates strongly, at $r=+0.681$ with one-tailed $p=0.031$ (left panel of \cref{fig:pa-vs-vqa-simpler}).

\begin{figure}[!htbp]

  \centering
  \includegraphics[height=0.245\textwidth]{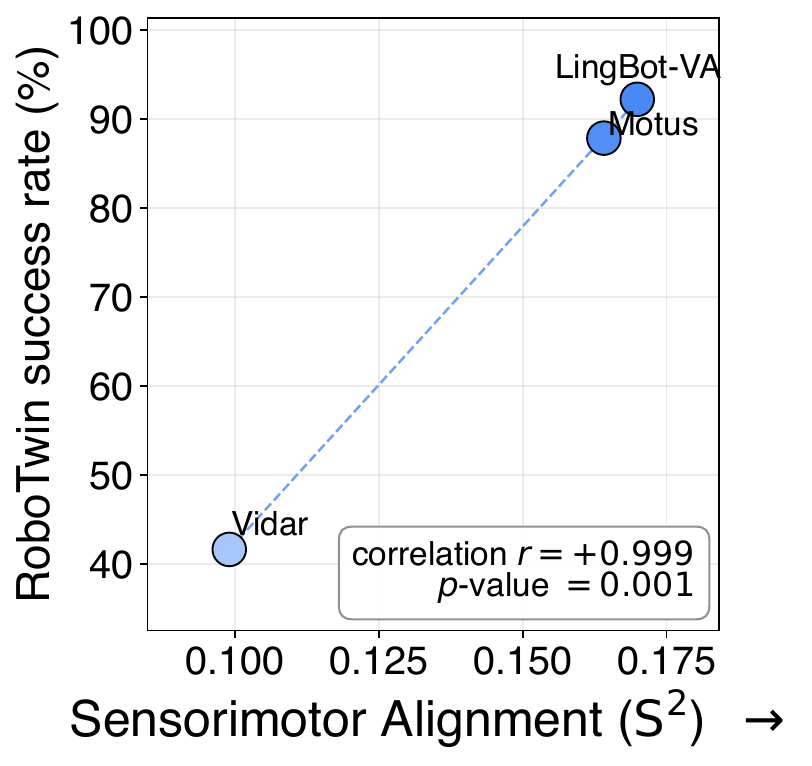}
  \hfill
  \includegraphics[height=0.245\textwidth]{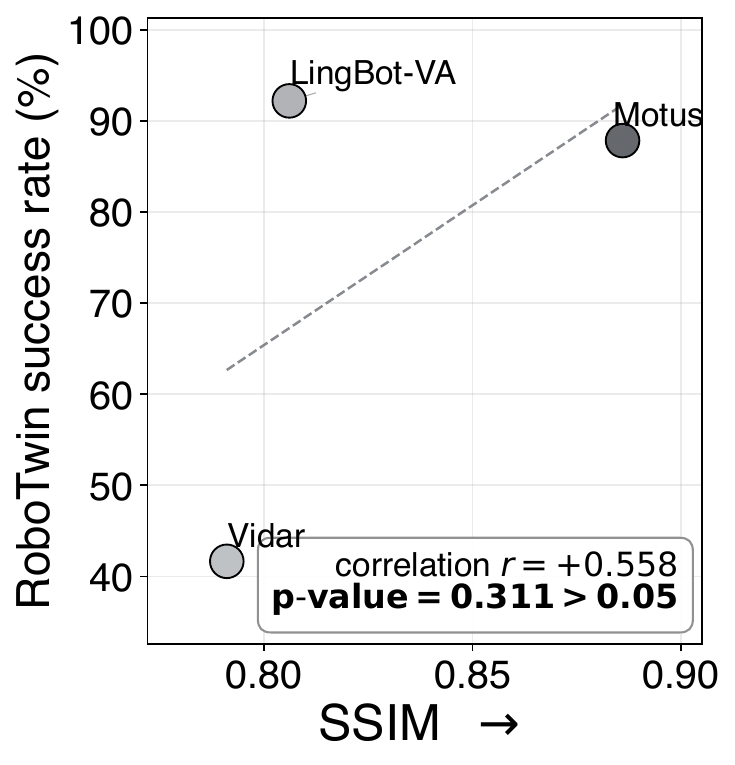}
  \hfill
  \includegraphics[height=0.245\textwidth]{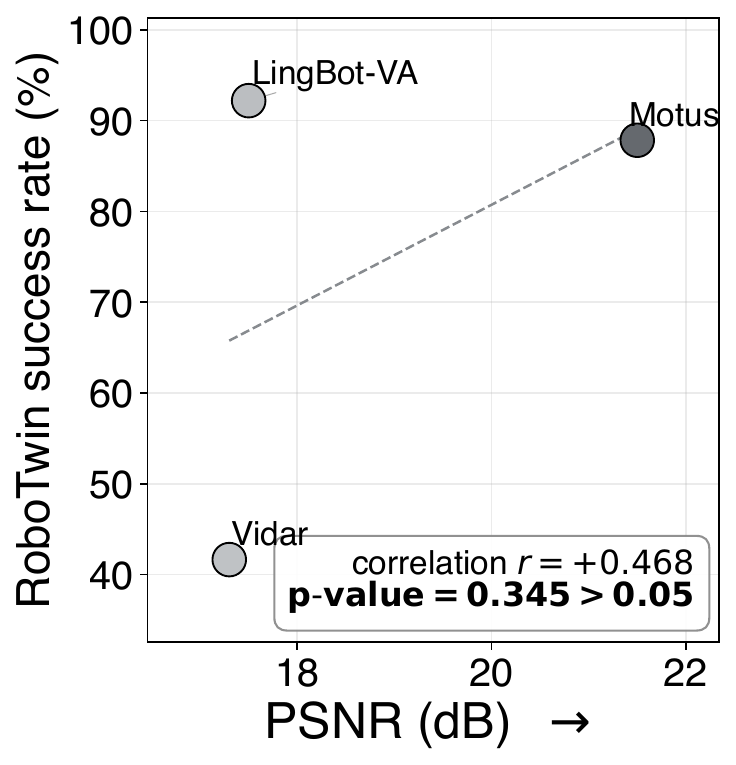}
  \hfill
  \includegraphics[height=0.245\textwidth]{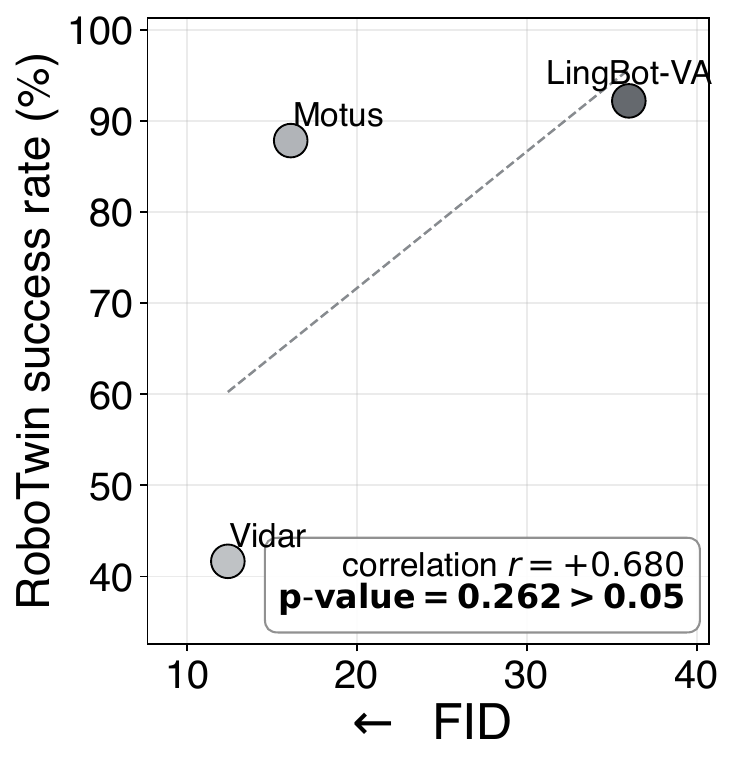}
  \caption{\textbf{Sensorimotor Alignment (\PA) correlates with WAM success rate more closely than video-quality metrics.}
For 3 world action models (WAMs) evaluated on RoboTwin, video-quality metrics (SSIM, PSNR, FID) show no significant correlations with task success.
In contrast, only \PA correctly recovers the success-rate ordering.
Only 3 of the 6 WAMs are compared because the others do not generate future video during closed-loop task evaluation.\label{fig:pa-vs-video-quality}}
\end{figure}

\textbf{\PA correlates with WAM success rate more closely than video-quality metrics.}
We compare \mbox{\PA} with standard video-quality metrics (FID~\citep{heusel2017gans}, SSIM~\citep{wang2004image}, and PSNR~\citep{hore2010image}) on 3 of the WAMs, Motus~\citep{bi2025motus}, LingBot-VA~\citep{li2026causal}, and Vidar~\citep{feng2025vidar}, evaluated on RoboTwin.
We find that these metrics correlate weakly with task success: SSIM reaches only \mbox{$r=+0.558$} and PSNR only \mbox{$r=+0.468$}, while FID correlates in the wrong direction, at \mbox{$r=+0.680$}, associating worse video quality with higher success (\mbox{\cref{fig:pa-vs-video-quality}}). By contrast, \mbox{\PA} on ALOHA correlates more strongly, at $r=+0.999$.

\FloatBarrier

\subsection{\PA Is Robust to Evaluation Choices}
\label{sec:finding-robustness}

We next check that the correlation does not depend on a particular choice of settings. We test this in two ways. First, we vary how \PA is computed and find that the positive correlation with task success holds. Second, as a negative control, we deliberately compute \PA on reference data from a mismatched embodiment, where a sound metric should lose this correlation. The correlation drops as expected, validating the design of \PA.

\begin{figure}[!htbp]
  \captionsetup{justification=raggedright}
  \centering
  \begin{minipage}{0.62\textwidth}
    \includegraphics[width=\linewidth]{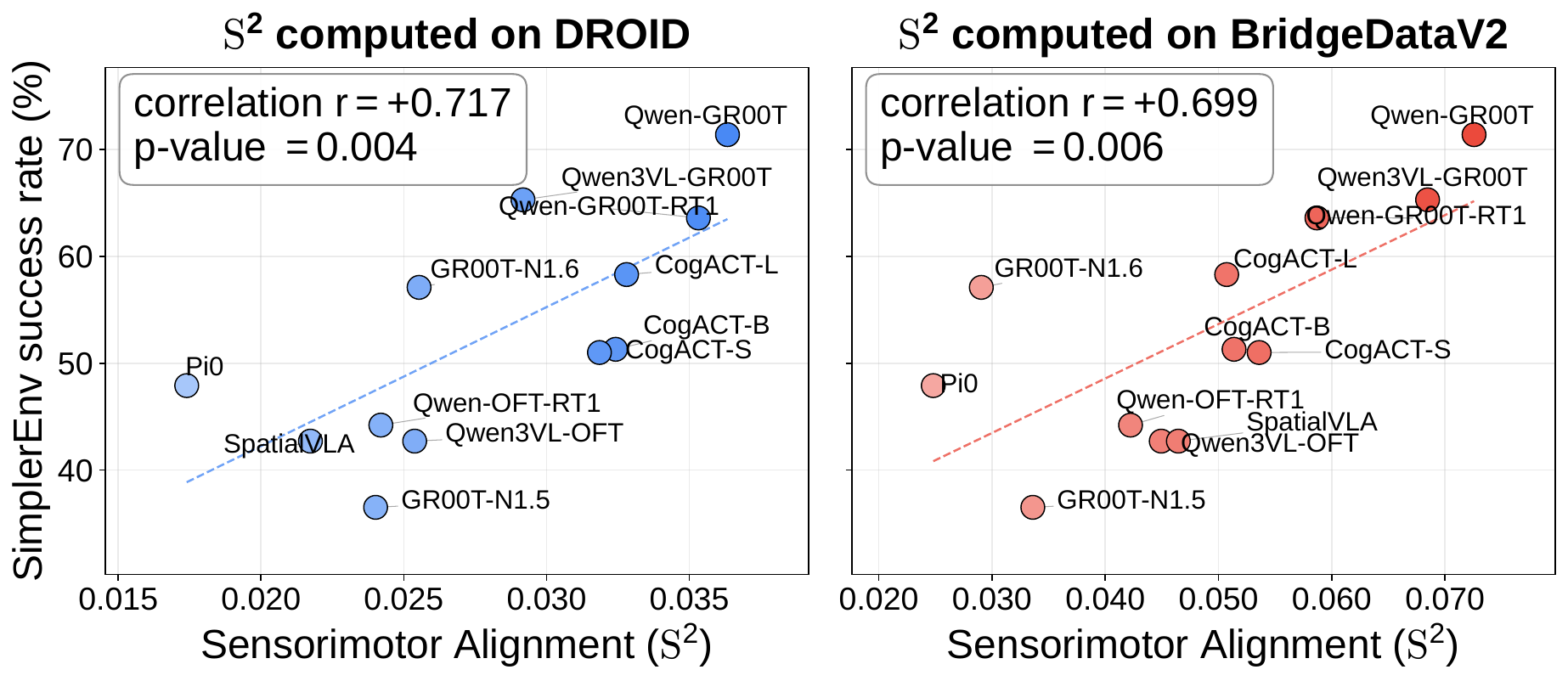}
  \end{minipage}\hfill
  \begin{minipage}{0.35\textwidth}
    \includegraphics[width=\linewidth]{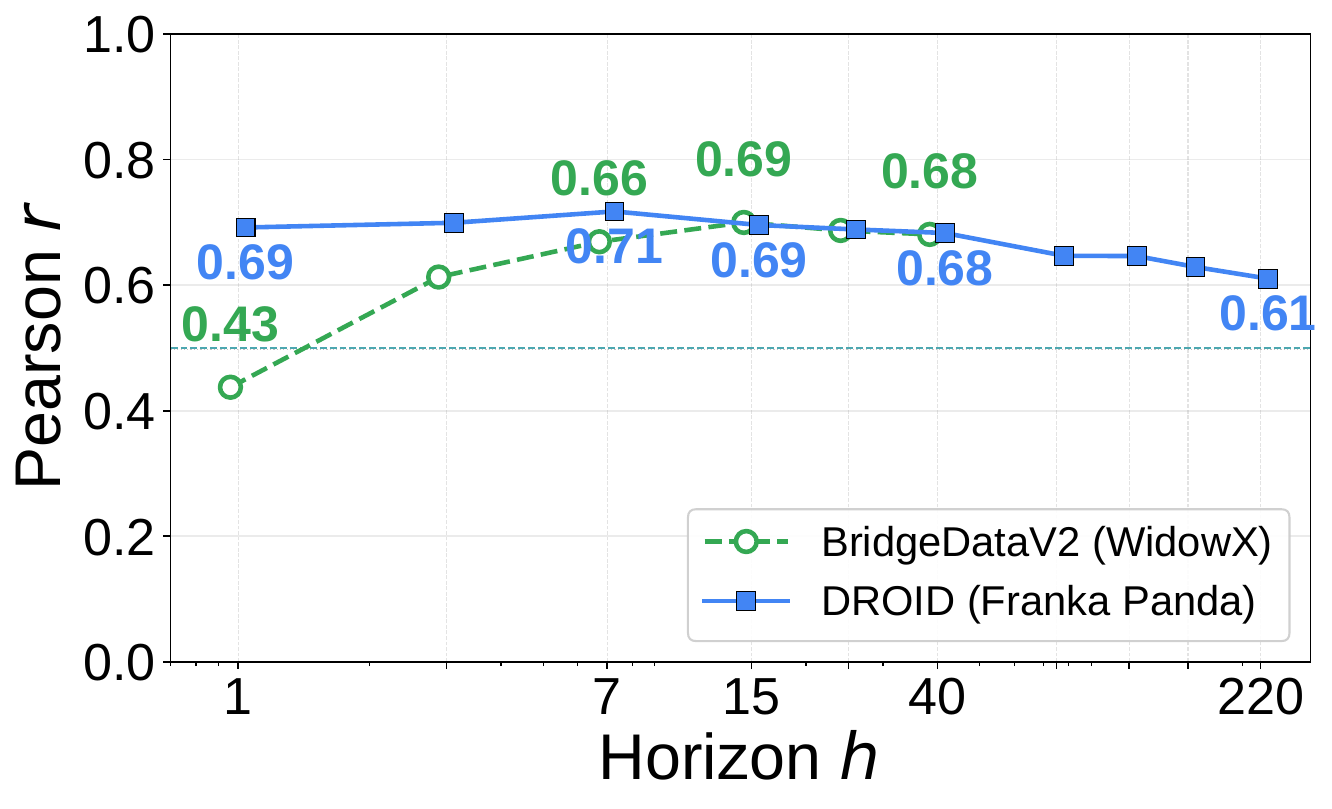}
  \end{minipage}
  \par
  \begin{minipage}[t]{0.55\textwidth}
    \caption{\textbf{The correlation transfers across reference datasets.}
For 12 vision-language-action models, Sensorimotor Alignment (\PA) on DROID (Franka Panda, left) and BridgeDataV2 (WidowX, right) correlates strongly with the same SimplerEnv-WidowX task success rates.\label{fig:cross-embodiment-vla}}
  \end{minipage}\hfill
  \begin{minipage}[t]{0.42\textwidth}
    \caption{\textbf{Stable correlations across horizons.}
For 12 vision-language-action models, Sensorimotor Alignment (\PA) and SimplerEnv-WidowX success rate have Pearson $r>0.5$ at nearly all tested horizons $h$ (future steps) on DROID and BridgeDataV2.}
    \label{fig:horizon-vla-llmout-vt}
  \end{minipage}
\end{figure}

\textbf{\PA is stable across computation choices.}
In the first check, we vary eight settings: the reference dataset, the trajectory horizon, five implementation choices, and DTW warping. The correlation stays positive under all eight. We discuss the first two below; Appendix~\ref{app:cknna-sr-full} gives per-model \PA by horizon, and Appendices~\ref{app:implementation-sensitivity} and~\ref{app:dtw-ablation} cover the other six.

The correlation transfers across reference datasets. We compute \PA on both DROID (Franka Panda) and BridgeDataV2~\citep{walke2023bridgedata} (WidowX), robots with different arm kinematics, and both correlate near $r\!=\!{+}0.7$ with task success for the 12 VLAs on SimplerEnv (\cref{fig:cross-embodiment-vla}).

The correlation is also stable across trajectory horizons. It stays above $r\!=\!0.5$ at all DROID horizons through $h\!=\!220$ and at BridgeDataV2 horizons from $h\!=\!3$ through $40$ (\cref{fig:horizon-vla-llmout-vt}).

\textbf{\PA is sensitive to embodiment mismatch.}
In the negative control, we compute \PA for 3 bimanual WAMs on a single-arm reference dataset (DROID) rather than the matched bimanual one (ALOHA). This embodiment mismatch drops the \PA--success correlation from $r\!=\!{+}0.999$ to $r\!=\!{+}0.246$, indicating that \PA responds to the difference in arm configuration rather than reporting alignment indiscriminately. Appendix~\ref{app:embodiment-mismatch} gives the full comparison.

\FloatBarrier

\section{Representation-Level Explanations with \texorpdfstring{\PA}{S2}}

\label{sec:pa-reveals}

Several phenomena reported for VLAs and WAMs rest only on closed-loop results and still lack a representation-level explanation. Because \mbox{\PA} correlates with task success, as shown in \cref{sec:pa-predicts-sr}, we use it to explain each of these phenomena, as summarized below:

\begin{itemize}[leftmargin=*,itemsep=0mm]

\item
\textbf{Modality dominance}:
    prior work observes that VLAs often struggle with instruction grounding: they act on what the scene visually suggests rather than on what the language instruction asks~\citep{libero-plus,libero-cf,igar}.
    In \cref{sec:finding-what-carries}, we show that after VLA fine-tuning, task success correlates with vision-feature \mbox{\PA} but not text-feature \mbox{\PA}, indicating that fine-tuned VLAs draw on what they see, not on the instruction, to produce their motion.
\item
\textbf{Frozen encoders}:
    prior work observes that freezing the vision encoder during VLA fine-tuning degrades downstream task performance~\citep{vlm4vla}.
    In \cref{sec:pa-reveals-frozen}, we show that freezing also prevents the \mbox{\PA} gain from end-to-end fine-tuning, indicating that VLAs gain performance by bringing their sensory representation closer to the motor representation.
\item
\textbf{Iterative denoising}:
    prior work observes that WAMs can skip iterative video denoising at test time, decoding actions without generating future video, with little loss in task success~\citep{yuan2026fastwam}.
    In \cref{sec:pa-reveals-denoising}, we show that the \mbox{\PA} of WAM features is already high at the earliest denoising step and changes little afterwards, indicating that the first denoising steps already yield features aligned closely enough with robot trajectories to act on.
\item
\textbf{Mid-depth blocks}:
    prior work observes that mid-depth blocks of video diffusion transformers matter most for semantic reasoning~\citep{wang2026demystifing}.
    In \cref{sec:pa-reveals-midlayer}, we show that these blocks also carry the strongest \mbox{\PA}, indicating that the layers that reason about video semantics are also the layers most aligned with robot motion.
\end{itemize}

\begin{figure}[!htbp]

  \centering
  \includegraphics[width=\textwidth]{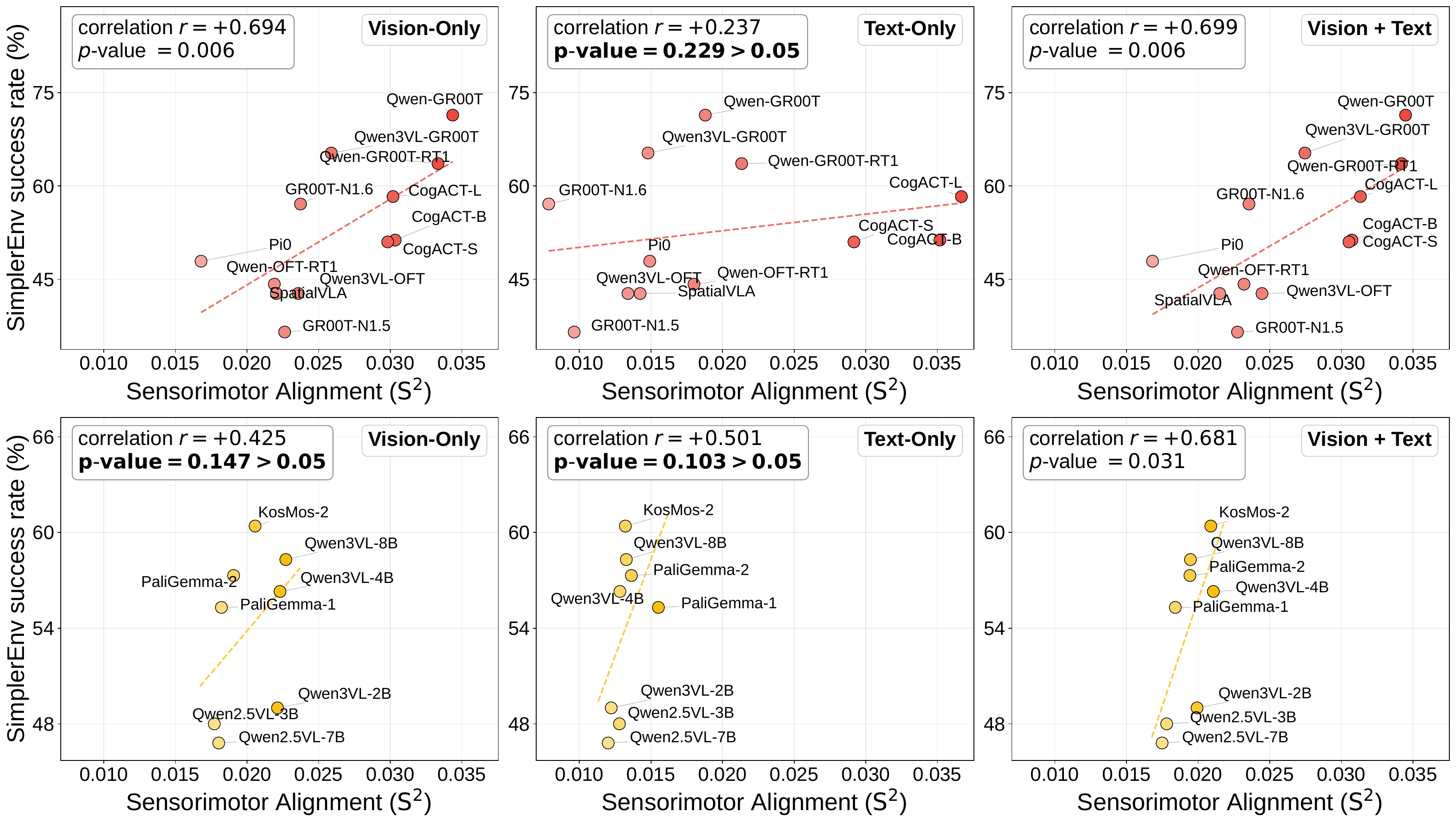}

\caption{\textbf{Fine-tuned VLAs rely mainly on visual rather than language representations.}
We correlate Sensorimotor Alignment (\PA) from vision, text, and joint vision-text features (left to right) with SimplerEnv task success rate.
\textbf{Top:} For fine-tuned vision-language-action models (VLAs), vision and joint features correlate similarly strongly, while text features correlate weakly.
\textbf{Bottom:} For raw vision-language models (VLMs), joint-feature \PA correlates with task success after VLA fine-tuning, while neither modality alone reaches significance.}
\label{fig:vt-what-carries-2x3}

\end{figure}

\subsection{What Carries the \texorpdfstring{Correlation}{Correlation}: Vision, Text, or Both?}
\label{sec:finding-what-carries}
VLAs often struggle with instruction grounding~\citep{libero-plus,libero-cf,igar}. We use \mbox{\PA} to explain this at the representation level, asking which modality the correlation with task success depends on: vision, text, or both. In \cref{fig:vt-what-carries-2x3}, we compute \mbox{\PA} separately for each modality. We take the VLM output tokens read by the action head and form \emph{vision features} from the image tokens, \emph{text features} from the instruction tokens, and \emph{joint vision-text features} from all tokens.

\textbf{Fine-tuned VLAs rely mainly on visual rather than language representations.}
In the 12 fine-tuned VLAs, \mbox{\PA} from vision features correlates with SimplerEnv success rate at \mbox{$r\!=\!{+}0.694$}, nearly matching the joint vision-text features at \mbox{$r\!=\!{+}0.699$}, whereas text features show no significant correlation at \mbox{$r\!=\!{+}0.237$}. The raw VLMs reverse this pattern: before fine-tuning, text features correlate at \mbox{$r\!=\!{+}0.501$}, more strongly than vision features at \mbox{$r\!=\!{+}0.425$}, and joint vision-text features correlate most strongly at \mbox{$r\!=\!{+}0.681$}.

We infer that, after fine-tuning, VLAs produce their motion mainly from visual rather than language representations, which explains their weak instruction grounding. Appendix~\ref{app:full-cknna-per-model} reports per-model \mbox{\PA} at all feature positions.

\subsection{Freezing Vision Encoders Limits \PA Gain}
\label{sec:pa-reveals-frozen}

\begin{figure}[!htbp]

  \captionsetup{justification=raggedright}
  \centering
  \begin{minipage}{0.55\textwidth}

    \includegraphics[width=\linewidth]{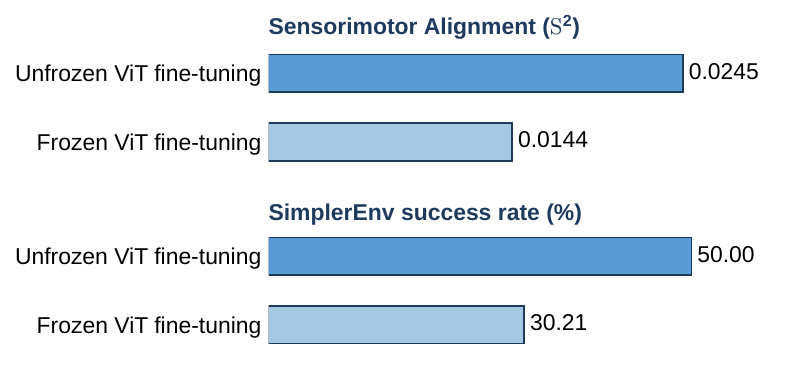}
  \end{minipage}\hfill
  \begin{minipage}{0.4\textwidth}

\caption{\textbf{Freezing the vision encoder reduces both Sensorimotor Alignment (\PA) and task success rate.}
For Qwen2.5-VL-3B vision-language-action models fine-tuned on BridgeDataV2, we compare end-to-end training with a frozen vision encoder.
Freezing lowers \PA from joint vision-text features (top) and SimplerEnv success rate (bottom), linking lower success to weaker alignment.
Trajectory horizon $h=7$.}
\label{fig:freeze-vs-unfreeze-bars}
\end{minipage}

\end{figure}

VLM4VLA~\citep{vlm4vla} reports that VLAs perform substantially worse when their vision encoder stays frozen during fine-tuning. We use \mbox{\PA} to explain this drop at the representation level, testing whether freezing also prevents the sensorimotor-alignment gain from end-to-end fine-tuning.

In \cref{fig:freeze-vs-unfreeze-bars}, we compare 2 VLAs that follow the same VLM4VLA~\citep{vlm4vla} recipe and differ only in whether the vision encoder is trained or kept frozen. Freezing the encoder reduces \PA by $41\%$ and SimplerEnv success rate by $40\%$. These results suggest that end-to-end fine-tuning improves performance by aligning the sensory representation more closely with the motor representation. Appendix~\ref{app:freeze-ablation-horizon-sweep} gives the full horizon sweep.

\FloatBarrier

\subsection{Features at Early Denoising Steps Already Carry Strong \ourmethod}
\label{sec:pa-reveals-denoising}

\begingroup
\setlength{\intextsep}{2pt}
\begin{figure}[!htbp]

  \captionsetup[subfigure]{skip=2pt}
  \centering
  \subcaptionbox{\PA is nearly flat across the schedule.\label{fig:pa-vs-denoising-pct}}[0.49\textwidth]{%
    \includegraphics[width=\linewidth]{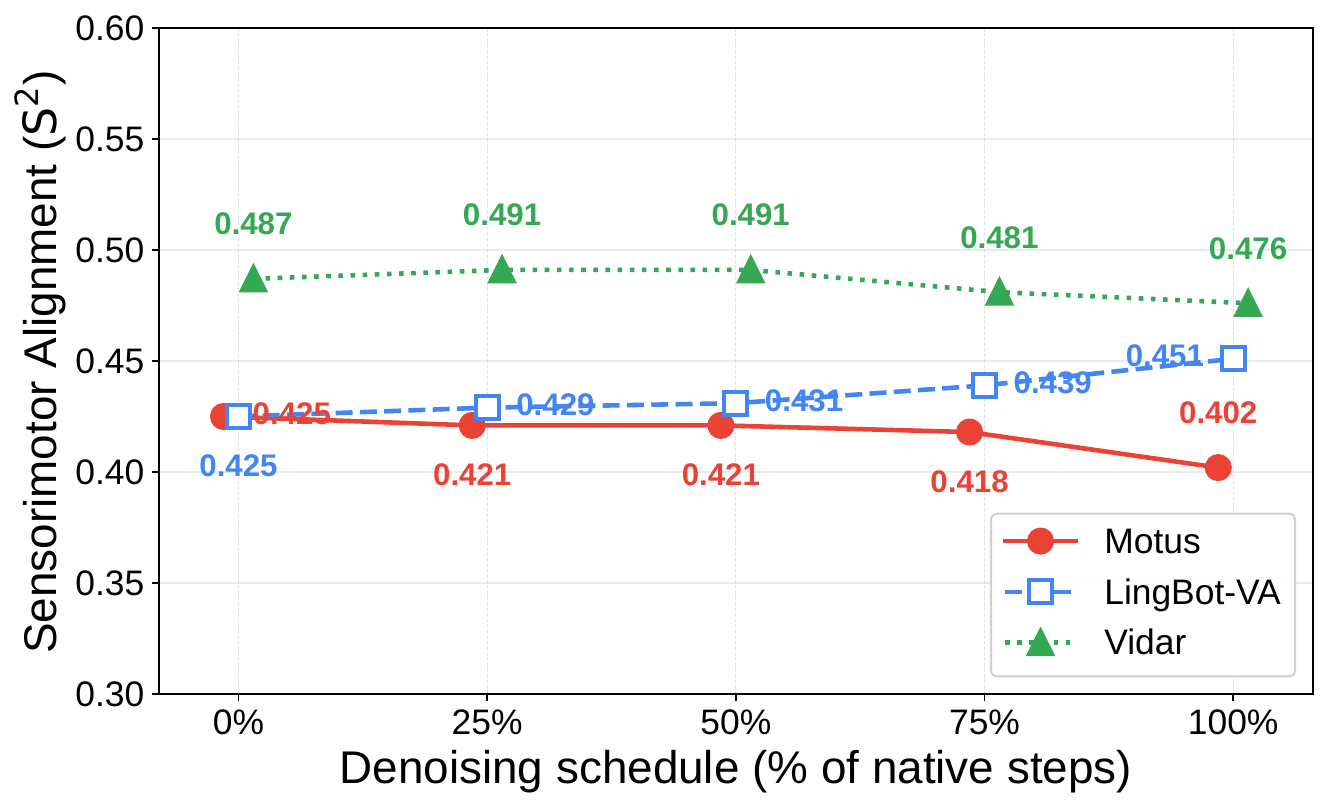}}
  \hfill
  \subcaptionbox{Block~14 has the strongest \PA in all 3 WAMs.\label{fig:pa-layerwise}}[0.49\textwidth]{%
    \includegraphics[width=\linewidth]{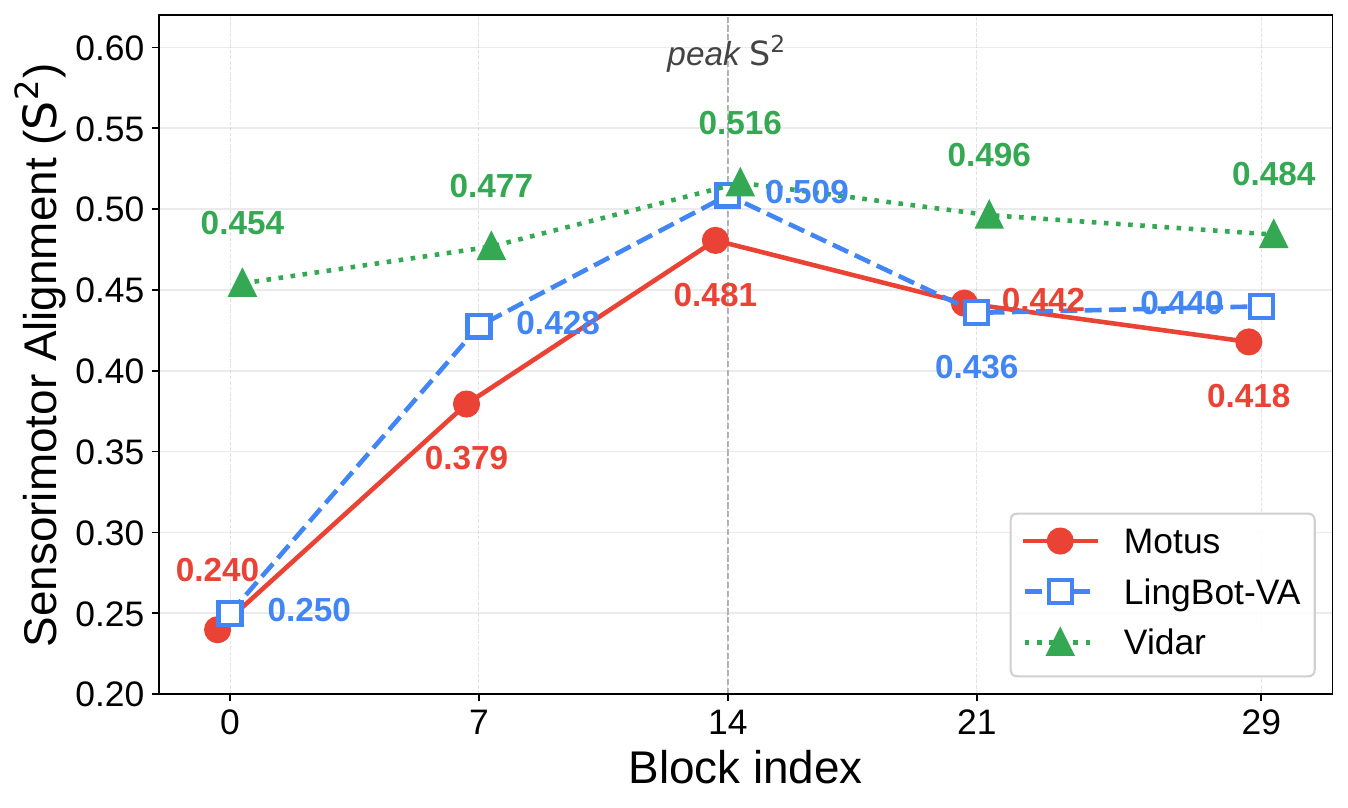}}

  \caption{\textbf{Sensorimotor Alignment (\PA) is already high at early denoising steps and peaks at mid-depth blocks.}
For 3 world action models (WAMs), \textbf{(a)} \PA remains nearly flat across denoising, suggesting early-step features already align with robot trajectories; \textbf{(b)} schedule-averaged \PA peaks at Block~14, in the middle of the 30-block backbone.}

\end{figure}
\endgroup

Most WAMs generate future video through iterative denoising and decode actions from the resulting latents. Fast-WAM~\citep{yuan2026fastwam} shows that WAMs can skip this video generation and decode actions directly, with little loss in task success. We examine the representation-level reason: whether features at the earliest denoising steps already align with robot trajectories.

We measure \mbox{\PA} for 3 WAMs~\citep{bi2025motus,li2026causal,feng2025vidar} on RoboTwin~\citep{mu2025robotwin} at 5 positions in each model's native denoising schedule, from \mbox{$0\%$} to \mbox{$100\%$}, not at matched noise levels (\cref{fig:pa-vs-denoising-pct}). \mbox{\PA} is already high at the earliest step and remains nearly flat, varying by about \mbox{$6\%$} at most from its initial value across the 3 models. We infer that early-step features are already sufficiently aligned with robot trajectories to generate actions. Because \mbox{\PA} correlates with WAM success rate (\cref{sec:finding-va-sr}), this result explains why skipping iterative video denoising barely lowers task success.

\subsection{Mid-Depth Blocks Carry the Strongest \ourmethod}
\label{sec:pa-reveals-midlayer}

Chain-of-Steps~\citep{wang2026demystifing} shows that mid-depth layers in Video Diffusion Transformers are especially important for semantic reasoning over objects and interactions. We ask whether the layers with the highest \mbox{\PA} also lie in this mid-depth range.

In \cref{fig:pa-layerwise}, we compute \mbox{\PA} at different depths of 3 WAMs~\citep{bi2025motus,li2026causal,feng2025vidar} on RoboTwin. All 3 share a 30-block video backbone; we extract features from Blocks 0, 7, 14, 21, and 29. Block~14 yields the highest \mbox{\PA} in every model. This peak falls in the mid-depth range that Chain-of-Steps~\citep{wang2026demystifing} identifies as semantically important, suggesting that the layers handling video semantic reasoning align most closely with robot motion. Details are in Appendix~\ref{app:layerwise-vs-denoising-step}.

\FloatBarrier
\section{Related Work}
\label{sec:related}

\textbf{Representation learning in robotics.}
Many robotics studies pretrain visual representations to improve task performance~\citep{r3m,mvp,vc1}. However, there remains limited understanding of which representational properties lead to downstream task success. A few studies ask what makes a pretrained vision encoder useful for robot learning. \citet{hu2023policy} evaluate 14 frozen vision encoders and find that the correlation between an encoder's offline score and downstream task success rate is not stable across policy-learning methods. \citet{burns2023makes} find that emergent segmentation predicts manipulation success under visual distribution shifts, but only for Vision Transformers. Theia~\citep{shang2024theia} hypothesizes that higher feature-norm entropy improves robot learning, but reports the supporting correlations only within separate groups of encoders. These studies examine only vision encoders rather than a broader range of embodied models, and none of them accounts for the trajectories the robot executes when measuring the representation. By contrast, \mbox{\PA} compares $k$-nearest-neighbor sets in sensory and motor spaces. The same \mbox{\PA} measure correlates with task success across VLAs, raw VLMs, WAMs, and the encoders that \citet{hu2023policy} evaluate.

\textbf{Platonic Representation Hypothesis.}
Recent work has applied the Platonic Representation Hypothesis (PRH)~\citep{huh2024platonic} to several new domains. In image generation, REPA~\citep{yu2024repa} aligns diffusion-transformer features with self-supervised vision encoders, improving training speed and quality. CREPA~\citep{hwang2025cross} extends this recipe to video diffusion. In multimodal understanding, VIRAL~\citep{yoon2025visual} re-aligns multimodal large language models (LLMs) with their vision encoders, recovering visual detail lost in multimodal training. For cross-modal alignment, \citet{schnaus2026shared} match independently trained vision and language encoders through their shared geometry, without paired data. PRH has not yet been brought into robotics; we extend it to robotics by treating robot trajectories as another projection of the world, and find that more capable embodied models align their sensory representations more closely with it.

\section{Conclusion}
\label{sec:conclusion}

We introduce \ourmethod (\PA), a training-free offline metric of alignment between sensory representations and end-effector trajectories. Across VLAs, raw VLMs, and WAMs in simulation, and across real-robot policies, \mbox{\PA} correlates positively with downstream task success, supporting a PRH-inspired view of embodiment. \mbox{\PA} correlates more strongly with task success than VQA scores and video-quality metrics, and explains, at the representation level, VLA and WAM phenomena previously observed only through closed-loop results: modality dominance, the performance drop from freezing the vision encoder, the small task-success cost of skipping iterative denoising, and the importance of mid-depth blocks for semantic reasoning. We use \mbox{\PA} only offline as a low-cost model-selection and diagnostic tool before closed-loop evaluation, but this correlation motivates optimizing embodied models directly for sensorimotor alignment.

\begin{samepage}
\textbf{Limitations.}
Our study has three main limitations. (1) The positive relationship between \mbox{\PA} and task success is correlational across the tested models; we have not yet demonstrated that sensorimotor alignment can be used directly during training to improve task success rates. (2) Models with similar \mbox{\PA} values can still achieve different success rates, so \mbox{\PA} is best used as a low-cost offline proxy for identifying promising models and VLM backbones before costly fine-tuning or closed-loop testing, which remains the final step of model selection. (3) Our evaluation covers limited model families, benchmarks, and training recipes, few models per comparison, and a single real-robot setting; broader generalization remains to be tested.

\end{samepage}

\bibliographystyle{preprint}
\bibliography{main}

\appendix

\clearpage
\section*{Appendix}
\noindent\textbf{Contents}

\noindent \hyperref[app:exp-setup]{A. Sensorimotor Alignment Metric}\dotfill\pageref{app:exp-setup}

\hyperref[app:exp-datasets]{\hspace{1em}A.1. Reference Datasets and Sampling}\dotfill\pageref{app:exp-datasets}

\hyperref[app:base-frame]{\hspace{1em}A.2. Trajectory Distance}\dotfill\pageref{app:base-frame}

\hyperref[app:cknna-estimator]{\hspace{1em}A.3. Alignment Estimator}\dotfill\pageref{app:cknna-estimator}

\noindent \hyperref[app:more-success]{B. Experiments on S$^2$ and Task Success}\dotfill\pageref{app:more-success}

\hyperref[app:vla-protocol]{\hspace{1em}B.1. Vision-Language-Action Models}\dotfill\pageref{app:vla-protocol}

\hyperref[app:confound-controls]{\hspace{1em}B.2. Confound Controls for the VLA Correlation}\dotfill\pageref{app:confound-controls}

\hyperref[app:models]{\hspace{1em}B.3. Raw Vision-Language Models}\dotfill\pageref{app:models}

\hyperref[app:wam-protocol]{\hspace{1em}B.4. World Action Models}\dotfill\pageref{app:wam-protocol}

\hyperref[app:real-robot-protocol]{\hspace{1em}B.5. Real-Robot Policies}\dotfill\pageref{app:real-robot-protocol}

\hyperref[app:hu-backbones]{\hspace{1em}B.6. Frozen Vision Encoders Under Downstream Policy Learning}\dotfill\pageref{app:hu-backbones}

\noindent \hyperref[app:robustness]{C. Robustness of S$^2$}\dotfill\pageref{app:robustness}

\hyperref[app:cknna-sr-full]{\hspace{1em}C.1. Trajectory Horizon}\dotfill\pageref{app:cknna-sr-full}

\hyperref[app:implementation-sensitivity]{\hspace{1em}C.2. Implementation Choices}\dotfill\pageref{app:implementation-sensitivity}

\hyperref[app:dtw-ablation]{\hspace{1em}C.3. DTW Alignment}\dotfill\pageref{app:dtw-ablation}

\hyperref[app:embodiment-mismatch]{\hspace{1em}C.4. Sensitivity to Embodiment Mismatch}\dotfill\pageref{app:embodiment-mismatch}

\noindent \hyperref[app:diagnostics]{D. Details of the Representation-Level Explanations}\dotfill\pageref{app:diagnostics}

\hyperref[app:full-cknna-per-model]{\hspace{1em}D.1. S$^2$ Across Feature Positions}\dotfill\pageref{app:full-cknna-per-model}

\hyperref[app:freeze-ablation-horizon-sweep]{\hspace{1em}D.2. Frozen Versus Unfrozen Vision Encoder}\dotfill\pageref{app:freeze-ablation-horizon-sweep}

\hyperref[app:layerwise-vs-denoising-step]{\hspace{1em}D.3. WAM Features Along the Denoising Schedule}\dotfill\pageref{app:layerwise-vs-denoising-step}

\clearpage
\section{Sensorimotor Alignment Metric}
\label{app:exp-setup}

This section gives the full computation of \mbox{\PA} summarized in \cref{sec:pa}: the reference datasets and how we sample from them, the trajectory distance that defines motor neighbors, and the estimator that scores alignment between sensory and motor neighbor sets.

\subsection{Reference Datasets and Sampling}
\label{app:exp-datasets}

We use three reference datasets: the single-arm DROID and BridgeDataV2 datasets for VLAs, raw VLMs, frozen vision encoders, and real-robot policies, and the bimanual ALOHA dataset for WAMs. DROID and BridgeDataV2 contain teleoperated robot demonstrations with RGB images, proprioceptive states, and language task annotations. Each state is a 7D vector $\mathbf{s}=(x,y,z,\phi,\theta,\psi,g)$, where $(x,y,z)$ is the end-effector position in meters, $(\phi,\theta,\psi)$ are the roll, pitch, and yaw angles in radians, and $g$ is the recorded gripper position. This shared format supports the same DTW and \mbox{\PA} computation on both single-arm datasets.

\textbf{BridgeDataV2.}
BridgeDataV2~\citep{walke2023bridgedata} contains 53{,}192 WidowX episodes recorded at 5\,Hz. We use the recorded 7D state and retain 38{,}307 episodes with non-empty task labels.

\textbf{DROID.}
DROID~\citep{khazatsky2024droid} contains 95{,}658 episodes on a Franka Panda arm recorded at 15\,Hz, spanning 86 task categories across 564 scenes in offices, kitchens, and homes. We form the 7D state by concatenating the 6D Cartesian end-effector pose with the gripper position, which is $0$ when open and $1$ when closed. We retain episodes with a language annotation. As the observation image, we use the \mbox{\texttt{exterior\_image\_1\_left}} camera at \mbox{$180\!\times\!320$}, the fixed exterior view closest to the third-person camera of BridgeDataV2.

\textbf{ALOHA.}
For WAMs evaluated on bimanual RoboTwin, we sample $N=839$ trajectory windows from the bimanual ALOHA dataset~\citep{zhao2023learning}. Each trajectory row is the recorded 14-dimensional joint-position action vector, with six joints and a gripper coordinate per arm.

\textbf{Anchor sampling across horizons.}\label{app:fixed-samples}
We compute \mbox{\PA} at several trajectory horizons $h$. To keep these evaluations comparable, we draw one sample set at a maximum horizon $H_{\max}$ and reuse it at every shorter horizon. The sampling rule draws an integer anchor index $t_i$ for episode $i$ of length $L_i$ uniformly from $\{0,\ldots,L_i-1-H_{\max}\}$. At horizon $1\leq h\leq H_{\max}$, row $u$ of $\boldsymbol{\tau}_i$ is the recorded state at frame $t_i+u$, for $u=1,\ldots,h$. The image at frame $t_i$ and the task instruction form the fixed sensory input $\mathbf{x}_i$, whose embedding $\mathbf{z}_i$ is computed once. The horizon sweep therefore changes only the trajectory length, not the episodes, anchors, or sensory inputs.
On BridgeDataV2, we set $H_{\max}\!=\!40$. From a pool of 7,762 episodes with $L_i>40$, we retain non-empty task labels and at most three episodes per distinct task instruction, leaving $N=3{,}245$. We sweep $h\in\{1,3,7,15,25,40\}$. On DROID, we set $H_{\max}\!=\!220$, which covers about the same fraction of the dataset-wide mean episode length. The sampling rule also requires $L_i \geq 2H_{\max}\!+\!1\!=\!441$ so that anchors are not concentrated at the start of short episodes. For the \mbox{$N\!=\!6{,}000$} retained episodes, we sweep $h\!\in\!\{1, 3, 7, 15, 25, 40, 75, 110, 150, 220\}$.

\subsection{Trajectory Distance}
\label{app:base-frame}

For BridgeDataV2 and DROID, we compare end-effector trajectories in each robot's recorded base frame. We do not transform trajectories into a shared world or camera frame. Appendix~\ref{app:implementation-sensitivity} compares this choice with positions expressed relative to the start of each trajectory.

\textbf{Local cost.}
For the single-arm trajectories, let $\mathbf{p}_{i,u}=(x_{i,u},y_{i,u},z_{i,u})^{\top}$ and let $\mathbf{q}_{i,u}\in\mathbb{R}^{4}$ be the unit quaternion obtained from the recorded roll, pitch, and yaw using extrinsic rotations about the $x$, $y$, then $z$ axes. Quaternions $\mathbf{q}$ and $-\mathbf{q}$ represent the same rotation in $\SO(3)$, the group of three-dimensional rotations. For steps $u$ and $v$ of episodes $i$ and $j$, the local cost is
\[
\begin{aligned}
d_{\mathrm{pos}}&=\|\mathbf{p}_{i,u}-\mathbf{p}_{j,v}\|_2,\qquad
d_{\mathrm{grip}}=|g_{i,u}-g_{j,v}|,\\
d_{\mathrm{rot}}&=2\arccos\!\left(\min\{1,|\mathbf{q}_{i,u}^{\top}\mathbf{q}_{j,v}|\}\right),\\
c_{uv}&=w_{\mathrm{pos}}d_{\mathrm{pos}}+w_{\mathrm{rot}}d_{\mathrm{rot}}+w_{\mathrm{grip}}d_{\mathrm{grip}}.
\end{aligned}
\]
The pair $(i,j)$ is suppressed in $c_{uv}$ and the component distances. All weights default to one. Components use meters, radians, and recorded gripper values without further scaling.

\textbf{Dynamic time warping.} We use the symmetric2 DTW recurrence with no warping window. For a fixed trajectory pair, let $D_{uv}$ be the accumulated cost through steps $u,v$, with
\[
D_{11}=c_{11},\qquad D_{u1}=\sum_{a=1}^{u}c_{a1},\qquad
D_{1v}=\sum_{b=1}^{v}c_{1b}.
\]
For $u,v\geq2$,
\[
D_{uv}=\min\{D_{u-1,v}+c_{uv},\ D_{u,v-1}+c_{uv},\ D_{u-1,v-1}+2c_{uv}\}.
\]
For two $h$-step trajectories, the distance is $D_{hh}$ without path-length normalization; $h=1$ gives the single-state cost $c_{11}$. ALOHA trajectories (Appendix~\ref{app:exp-datasets}) use the same DTW recurrence with Euclidean distance between 14-dimensional joint-position vectors as the local cost.

\subsection{Alignment Estimator}
\label{app:cknna-estimator}

We follow the progression from HSIC to CKA and centered kernel nearest-neighbor alignment (CKNNA) in Appendix~A of \citet{huh2024platonic}, specializing their construction to a sensory kernel and a motor neighbor matrix. CKA scores alignment across all sample pairs; CKNNA restricts the comparison to mutual nearest neighbors, which measures alignment locally.

\textbf{Similarity matrices.}
For the $N$ paired samples in Section~\ref{sec:pa}, normalize sensory features as
\[
\widehat{\mathbf{z}}_i=\frac{\mathbf{z}_i}{\max(\|\mathbf{z}_i\|_2,\varepsilon_z)},
\qquad \varepsilon_z=10^{-12}.
\]
Let $\mathbf{K},\mathbf{L}\in\mathbb{R}^{N\times N}$ denote the sensory and motor matrices used in the score, with entries
\[
K_{ij}=\widehat{\mathbf{z}}_i^{\top}\widehat{\mathbf{z}}_j,
\qquad
L_{ij}=\mathbb{1}\{j\in\mathcal{N}_{\mathrm{motor}}(i)\}.
\]
Here $K_{ij}=\cos(\mathbf{z}_i,\mathbf{z}_j)$, with zero features mapped to zero, so $\mathbf{K}$ is the inner-product (linear) kernel on $\ell_2$-normalized features used by \citet{huh2024platonic}. The motor set $\mathcal{N}_{\mathrm{motor}}(i)$ contains the $k$ other samples with smallest DTW distance, as defined in Appendix~\ref{app:base-frame}. Thus DTW selects motor neighbors, and $\mathbf{L}$ assigns them unit weight; $\mathbf{L}$ is a directed similarity matrix, not necessarily a positive-semidefinite kernel. For $1\leq k<N$, we write $\mathcal{N}_{\mathbf{K}}(i)$ and $\mathcal{N}_{\mathbf{L}}(i)$ for the sets of $k$ other sample indices with the largest entries in row $i$ of $\mathbf{K}$ and $\mathbf{L}$; these are the sensory set $\mathcal{N}_{\mathrm{sensory}}(i)$ and motor set $\mathcal{N}_{\mathrm{motor}}(i)$ from Section~\ref{sec:pa}.

\textbf{HSIC and centered kernel alignment.}
For two kernel matrices $\mathbf{A},\mathbf{B}\in\mathbb{R}^{N\times N}$, let $\mathbf{H}=\mathbf{I}_N-\frac{1}{N}\mathbf{1}\mathbf{1}^{\top}$ be the centering matrix, where $\mathbf{I}_N$ is the identity and $\mathbf{1}\in\mathbb{R}^{N}$ is the all-ones vector. Define $\overline{\mathbf{A}}=\mathbf{H}\mathbf{A}\mathbf{H}$ and $\overline{\mathbf{B}}=\mathbf{H}\mathbf{B}\mathbf{H}$. The biased empirical Hilbert--Schmidt independence criterion (HSIC) and its normalized counterpart, centered kernel alignment (CKA)~\citep{kornblith2019similarity}, are
\[
\begin{aligned}
\mathsf{HSIC}_{\mathrm{b}}(\mathbf{A},\mathbf{B})
&=\frac{\operatorname{tr}(\overline{\mathbf{A}}\,\overline{\mathbf{B}})}{(N-1)^2},\\
\mathsf{CKA}(\mathbf{A},\mathbf{B})
&=\frac{\mathsf{HSIC}_{\mathrm{b}}(\mathbf{A},\mathbf{B})}
{\sqrt{\mathsf{HSIC}_{\mathrm{b}}(\mathbf{A},\mathbf{A})\,\mathsf{HSIC}_{\mathrm{b}}(\mathbf{B},\mathbf{B})}}.
\end{aligned}
\]
Here $\operatorname{tr}$ denotes the trace and $\top$ denotes transpose; the subscript $\mathrm{b}$ identifies the biased empirical estimator. Expanding the trace writes HSIC as a sum over all ordered sample pairs,
\[
\operatorname{tr}(\overline{\mathbf{A}}\,\overline{\mathbf{B}})=\sum_{i}\sum_{j}\overline{A}_{ij}\,\overline{B}_{ji},
\]
so CKA measures alignment globally, across every pair of samples, and is defined when both self terms are positive~\citep{huh2024platonic,kornblith2019similarity}. Restricting this sum to mutual nearest neighbors, as below, biases the measure toward local structure.

For nearest-neighbor alignment, we use the diagonal-corrected estimator of \citet{song2012feature}, as in the released CKNNA implementation of \citet{huh2024platonic}.\footnote{\url{https://github.com/minyoungg/platonic-rep/blob/main/metrics.py}} Let $\widetilde{\mathbf{A}},\widetilde{\mathbf{B}}$ be copies of $\mathbf{A},\mathbf{B}$ with zero diagonals. For $N>3$, define
\[
\widehat{\mathsf{HSIC}}_{\mathrm{u}}(\mathbf{A},\mathbf{B})
=\frac{1}{N(N-3)}\left[
\operatorname{tr}(\widetilde{\mathbf{A}}\widetilde{\mathbf{B}})
+\frac{(\mathbf{1}^{\top}\widetilde{\mathbf{A}}\mathbf{1})(\mathbf{1}^{\top}\widetilde{\mathbf{B}}\mathbf{1})}{(N-1)(N-2)}
-\frac{2\,\mathbf{1}^{\top}\widetilde{\mathbf{A}}\widetilde{\mathbf{B}}\mathbf{1}}{N-2}
\right].
\]
The subscript $\mathrm{u}$ denotes this unbiased estimator, which corrects the finite-sample bias of HSIC under i.i.d. sampling. We apply the same algebra to the directed neighbor matrices: the trace term is $\sum_{i,j}\widetilde{A}_{ij}\widetilde{B}_{ji}$, and we do not symmetrize the matrices.

\textbf{Centered kernel nearest-neighbor alignment.}
For $\mathbf{A},\mathbf{B}\in\{\mathbf{K},\mathbf{L}\}$, using the neighbor sets $\mathcal{N}_{\mathbf{K}}$ and $\mathcal{N}_{\mathbf{L}}$ defined above, we define the shared-neighbor mask for each matrix pair by
\[
\alpha_{ij}(\mathbf{A},\mathbf{B})
=\mathbb{1}\{j\in\mathcal{N}_{\mathbf{A}}(i)\cap\mathcal{N}_{\mathbf{B}}(i)\},
\]
where $\mathbb{1}\{\cdot\}$ is one when its condition holds and zero otherwise. Self-neighbors are excluded, and motor-neighbor indices are reused across models. The mask is directed: $\alpha_{ij}$ need not equal $\alpha_{ji}$. As in \citet{huh2024platonic}, ``mutual'' means membership in both neighbor sets.

Using $\odot$ for entrywise multiplication, define
\[
\mathsf{Align}_{\mathrm{knn}}(\mathbf{A},\mathbf{B})
=\widehat{\mathsf{HSIC}}_{\mathrm{u}}\!\left(
\boldsymbol{\alpha}(\mathbf{A},\mathbf{B})\odot\mathbf{A},
\boldsymbol{\alpha}(\mathbf{A},\mathbf{B})\odot\mathbf{B}\right).
\]
This follows the released CKNNA code's masking-before-HSIC computation. Separate masks cover the cross term $(\mathbf{K},\mathbf{L})$ and the self terms $(\mathbf{K},\mathbf{K})$ and $(\mathbf{L},\mathbf{L})$. Because $\mathbf{L}$ is binary, $\boldsymbol{\alpha}(\mathbf{K},\mathbf{L})\odot\mathbf{L}=\boldsymbol{\alpha}(\mathbf{K},\mathbf{L})$: shared neighbors retain sensory similarities and receive unit motor weights.

The reported score follows the CKNNA normalization:
\begin{equation}
\mathrm{S}^2=\frac{\mathsf{Align}_{\mathrm{knn}}(\mathbf{K},\mathbf{L})}
{\sqrt{\max\!\left\{
\mathsf{Align}_{\mathrm{knn}}(\mathbf{K},\mathbf{K})\,
\mathsf{Align}_{\mathrm{knn}}(\mathbf{L},\mathbf{L}),\varepsilon_A\right\}}},
\qquad \varepsilon_A=10^{-10}.
\label{eq:cknna-normalization}
\end{equation}
Following the relaxation of \citet{huh2024platonic}, treating all shared neighbors as equidistant replaces the sensory weights by one, and the resulting count of shared neighbors recovers the overlap of Equation~\ref{eq:neighbor-overlap}: $\sum_{i,j}\alpha_{ij}(\mathbf{K},\mathbf{L})=Nk\,m_{\mathrm{NN}}$, so $m_{\mathrm{NN}}=(Nk)^{-1}\sum_{i,j}\alpha_{ij}(\mathbf{K},\mathbf{L})\in[0,1]$. The directed, centered score $\mathrm{S}^2$ keeps the sensory weights and is not guaranteed to lie in this interval.

\section{Experiments on \texorpdfstring{\PA}{S2} and Task Success}
\label{app:more-success}

This section details the experiments in \cref{sec:finding-pa-predicts-across-families} by model family, plus VLA confound controls and a test of frozen vision encoders under downstream policy learning.

\subsection{Vision-Language-Action Models}
\label{app:vla-protocol}

The 12 evaluated VLAs include 5 StarVLA~\citep{community2026starvla} variants (Qwen2.5-GR00T-Bridge, Qwen2.5-GR00T-RT-1-Bridge, Qwen2.5-OFT-RT-1-Bridge, Qwen3-GR00T-RT-1-Bridge, and Qwen3-OFT-RT-1-Bridge), CogACT~\citep{li2024cogact} in Small, Base, and Large sizes, GR00T-N1.5-Bridge, GR00T-N1.6-Bridge~\citep{bjorck2025gr00t}, Pi0-Bridge~\citep{black2024pi_0}, and SpatialVLA-Bridge~\citep{qu2025spatialvla}. They cover 4 action-head types (flow matching, diffusion, autoregressive, and deterministic MLP) and 3 VLM training regimes, from frozen to end-to-end. We use reported SimplerEnv success rates from the model papers and public evaluation repositories, including \url{https://github.com/starVLA/starVLA} and \url{https://github.com/DelinQu/SimplerEnv-OpenVLA}.

\subsection{Confound Controls for the VLA Correlation}
\label{app:confound-controls}

The 12 VLAs in Appendix~\ref{app:vla-protocol} differ in fine-tuning recipe, model family and scale, and architecture group. Three checks show that none of these factors alone drives the correlation. Each check retains its main-paper experiment's reference data and feature settings.

\textbf{Training recipe.}
The raw-VLM experiment in Appendix~\ref{app:models} holds the recipe fixed. All 8 VLMs become VLAs under the same VLM4VLA~\citep{vlm4vla} recipe: the same action head, objective, BridgeDataV2 data, and hyperparameters. The correlation remains strong at $r\!=\!{+}0.681$ with one-tailed $p\!=\!0.031$, so the fine-tuning recipe does not drive the correlation.

\textbf{Model family and scale.}
CogACT~\citep{li2024cogact} Small, Base, and Large share the same VLM backbone and action-head design and differ only in action-head size. Within this family, the correlation is also positive, at $r\!=\!{+}0.826$. With only 3 models, this check has little statistical power (one-tailed $p\!=\!0.191$), but its direction matches the full cohort.

\textbf{Architecture groups.}
We remove each of the 5 architecture groups in turn and recompute the correlation on the remaining models, with results in \cref{tab:leave-one-out}. The correlation stays positive in all 5 cases, from $r\!=\!{+}0.526$ to $r\!=\!{+}0.785$, and remains significant at the $0.05$ level in 4 of them. The exception is removing the 5 StarVLA variants, which leaves 7 models with one-tailed $p=0.113$.

\begin{table}[!htbp]
  \centering
  \small
  \caption{\textbf{Leave-one-out over architecture groups} for 12 fine-tuned vision-language-action models.
Each row removes one group and recomputes the correlation between Sensorimotor Alignment (\PA) and SimplerEnv-WidowX task success rate.
The remaining model counts $n$ are 7, 9, 10, 11, and 11 in row order.\label{tab:leave-one-out}}
  \begin{tabular}{lcc}
    \toprule
    Removed group & Pearson's $r$ & One-tailed $p$ \\
    \midrule
    StarVLA    & $+0.526$ & $0.113$ \\
    CogACT     & $+0.785$ & $0.006$ \\
    GR00T      & $+0.745$ & $0.007$ \\
    Pi0        & $+0.783$ & $0.002$ \\
    SpatialVLA & $+0.687$ & $0.010$ \\
    \bottomrule
  \end{tabular}
\end{table}

\subsection{Raw Vision-Language Models}
\label{app:models}

We evaluate VLM4VLA's~\citep{vlm4vla} 8 raw VLMs: KosMos-2~\citep{peng2023kosmos}, PaliGemma~\citep{beyer2024paligemma}, PaliGemma 2~\citep{steiner2024paligemma}, Qwen2.5-VL-3B and Qwen2.5-VL-7B~\citep{bai2025qwen25vltechnicalreport}, and Qwen3-VL-2B, Qwen3-VL-4B, and Qwen3-VL-8B~\citep{bai2025qwen3}. Their success rates come from the VLAs that VLM4VLA fine-tunes from each backbone on BridgeDataV2 and evaluates on SimplerEnv. We compute \mbox{\PA} on the raw backbones before any robot fine-tuning, using DROID~\citep{khazatsky2024droid} as the reference dataset, which differs from the fine-tuning data in robot and scenes. We use $h=3$ and pool image and instruction tokens.

\subsection{World Action Models}
\label{app:wam-protocol}

We evaluate 6 WAMs: Vidar~\citep{feng2025vidar}, LingBot-VA~\citep{li2026causal}, Motus~\citep{bi2025motus}, Fast-WAM~\citep{yuan2026fastwam}, FlowWAM~\citep{chen2026flowwam}, and Kairos~\citep{kairos2026}. For all 6 models, we compute \mbox{\PA} on identical bimanual ALOHA trajectories, fixing feature position, trajectory horizon, number of neighbors, and DTW computation. The reported comparison uses video-backbone Block 7 features at $h=15$ and $k=10$.

\subsection{Real-Robot Policies}
\label{app:real-robot-protocol}

We use success rates from MolmoB0T's physical-robot evaluation~\citep{deshpande2026molmobot}, which tests all policies on the same tasks, robots, and cameras. Policies run on 3 Franka DROID platforms in 4 environments (kitchen, workroom, bedroom, and office) at 2 institutions. Each environment has 10 pick-and-place tasks with 3 trials per task, giving 120 trials per policy. A trial succeeds if the policy places the named object stably in or on the receptacle within 900 steps. The evaluation compares 5 policies, three MolmoB0T variants (MolmoB0T, MolmoB0T-Img, and MolmoB0T-Pi0) and two baselines (\mbox{$\pi_0$}-DROID and \mbox{$\pi_{0.5}$}-DROID). We compute \mbox{\PA} from public checkpoints' joint LLM-output features on the 6,000-episode DROID subset at $h=110$. This is an exploratory horizon; the prespecified $h=7$ gives $r=+0.429$ with one-tailed $p=0.236$.

\subsection{Frozen Vision Encoders Under Downstream Policy Learning}
\label{app:hu-backbones}

We ask whether \mbox{\PA} can rank frozen vision backbones without training a policy. \citet{hu2023policy} provide a benchmark: they evaluate 14 pretrained vision backbones (9 ResNet-50~\citep{he2016deep} and 5 ViT-B/16~\citep{dosovitskiy2021image}) on 21 tasks in Meta-World~\citep{yu2020metaworld}, Robosuite~\citep{zhu2020robosuite}, and Franka-Kitchen~\citep{gupta2019relay}. They freeze each backbone and train policies on top, so only the backbone changes. We compute \mbox{\PA} for their released checkpoints on the same 6,000-episode DROID subset as in the main paper, with all other settings unchanged. We take the interquartile mean (IQM) success rates from their paper for two policy-learning methods. Behavior cloning (BC) learns from expert demonstrations, and we use the smallest budget of 1 demonstration per task (5 on Robosuite). The visual reward function (VRF) method uses reinforcement learning, with rewards computed from distances between the backbone's embeddings of agent and expert observations.

\begin{figure}[!htbp]

  \centering
  \includegraphics[width=0.92\textwidth]{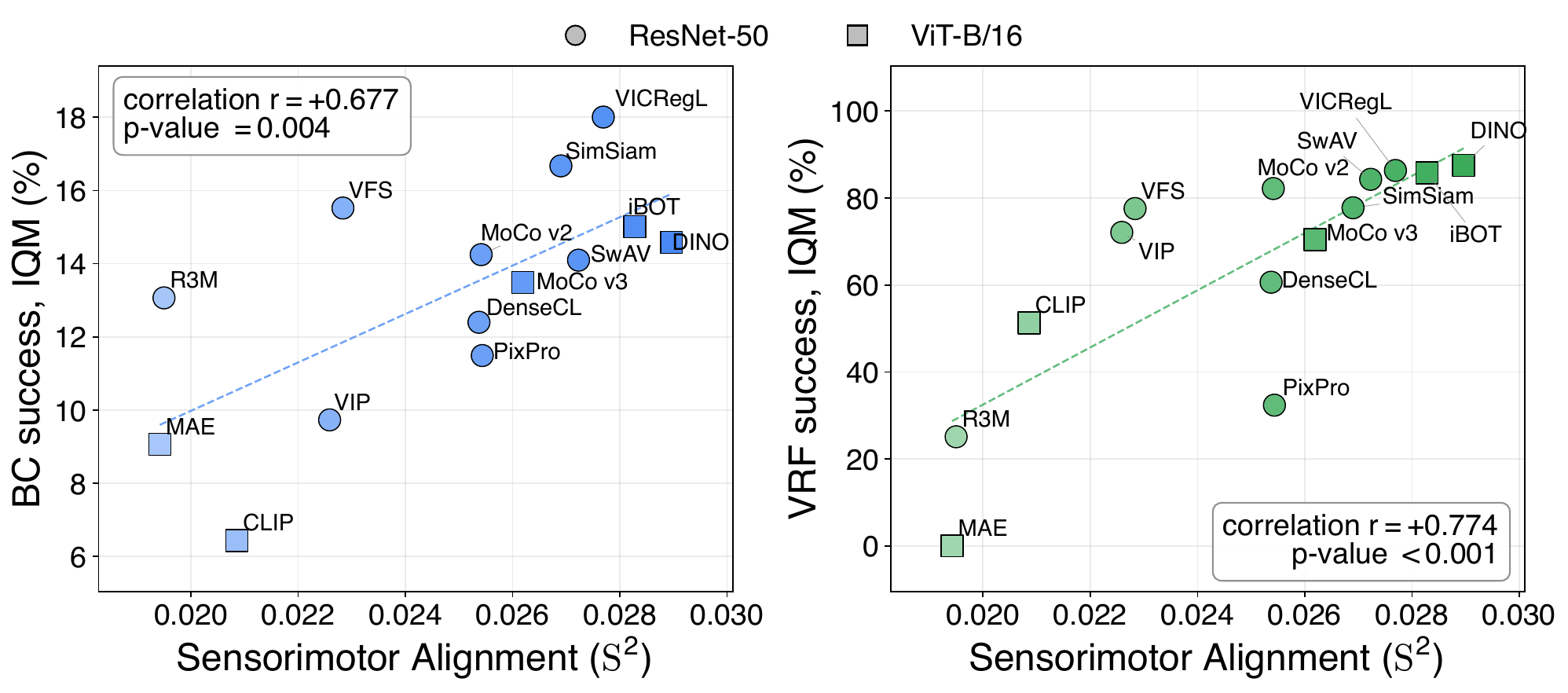}
  \caption{\textbf{Sensorimotor Alignment (\PA) of frozen vision encoders correlates with downstream policy success rate.}
Points represent the 14 frozen backbones of \citet{hu2023policy}, scored by \PA on DROID without policy training.
The vertical axis shows the interquartile mean (IQM) of task success rates over 21 tasks, using low-data behavior cloning (left) or reinforcement learning with visual reward functions (right).\label{fig:pa-vs-hu-backbones}}
\end{figure}

We find a strong correlation between the backbone's \mbox{\PA} and downstream task success under both policy-learning methods, as shown in \cref{fig:pa-vs-hu-backbones}. For BC, $r=+0.677$ with one-tailed $p=0.004$. For VRF, $r=+0.774$ with one-tailed $p<0.001$. Both methods rely directly on the frozen representation, and the correlation weakens when that reliance decreases. As the BC demonstration budget grows, the correlation drops to $r=+0.236$ at the largest budget, consistent with more demonstrations letting the policy compensate for a weaker representation. For reinforcement learning from task rewards, which \citet{hu2023policy} find too variable across random seeds to evaluate vision backbones reliably, we find no positive correlation on two Meta-World seed sets or on Robosuite, where $r$ lies between $-0.193$ and $-0.002$. Thus, \mbox{\PA} correlates with task success when the learning method relies directly on the frozen representation.

\FloatBarrier
\section{Robustness of \texorpdfstring{\PA}{S2}}
\label{app:robustness}

This section extends \cref{sec:finding-robustness}, which checks that the correlation does not depend on a particular choice of settings. Appendices~\ref{app:cknna-sr-full}--\ref{app:dtw-ablation} cover the first check, across the trajectory horizon, five implementation choices, and DTW warping. Appendix~\ref{app:embodiment-mismatch} gives the negative control, a mismatch in arm configuration.

\subsection{Trajectory Horizon}
\label{app:cknna-sr-full}

\cref{tab:cknna-sr-full-imgtext,tab:cknna-sr-full-img,tab:cknna-sr-full-txt} report \mbox{\PA} for the 12 finetuned VLAs at all 10 DROID horizons ($N\!=\!6{,}000$, $k\!=\!10$), for the joint, vision-token, and text-token features selected from the same LLM output; \cref{fig:horizon-vla-llmout-vt} summarizes the joint-feature correlation on both reference datasets. Models are sorted by SimplerEnv WidowX success rate (SR). Bottom rows give Spearman's $\rho$ and Pearson's $r$ with success rate, each followed by its one-tailed $p$-value for a positive association. We find that \mbox{\PA} from joint vision-text features and from vision features correlates significantly with the success rate at every horizon, with \mbox{$r$} from \mbox{${+}0.610$} to \mbox{${+}0.717$} and from \mbox{${+}0.554$} to \mbox{${+}0.703$}, respectively. Text-feature \mbox{\PA} is not significant under Pearson's \mbox{$r$} at any horizon, where \mbox{$r$} ranges from \mbox{${+}0.227$} to \mbox{${+}0.274$}, consistent with the modality analysis in \cref{sec:finding-what-carries}.

\begin{table}[p]
  \centering
  \scriptsize
  \setlength{\tabcolsep}{3pt}
  \caption{Sensorimotor Alignment (\PA) from joint vision-text features at the language-model output versus SimplerEnv-WidowX task success rate (SR) across trajectory horizons $h$.
We score 12 fine-tuned vision-language-action models on DROID ($N=6{,}000$ episodes, $k=10$ neighbors).
Bottom rows report Spearman's $\rho$ and Pearson's $r$ with their one-tailed $p$-values.
Bold marks the horizon of peak Pearson $r$ ($h=7$).}
  \label{tab:cknna-sr-full-imgtext}
  \begin{tabular}{lccccccccccc}
    \toprule
    Model & SR (\%) & $h\!=\!1$ & $h\!=\!3$ & $h\!=\!7$ & $h\!=\!15$ & $h\!=\!25$ & $h\!=\!40$ & $h\!=\!75$ & $h\!=\!110$ & $h\!=\!150$ & $h\!=\!220$ \\
    \midrule
    Qwen2.5-GR00T-Bridge      & 71.4 & 0.034 & 0.034 & \textbf{0.036} & 0.040 & 0.043 & 0.051 & 0.065 & 0.076 & 0.086 & 0.097 \\
    Qwen3-GR00T-RT-1-Bridge   & 65.3 & 0.028 & 0.027 & \textbf{0.029} & 0.032 & 0.036 & 0.044 & 0.057 & 0.071 & 0.083 & 0.096 \\
    Qwen2.5-GR00T-RT-1-Bridge & 63.6 & 0.034 & 0.034 & \textbf{0.035} & 0.040 & 0.042 & 0.048 & 0.062 & 0.071 & 0.080 & 0.090 \\
    CogACT-Large              & 58.3 & 0.031 & 0.031 & \textbf{0.033} & 0.037 & 0.040 & 0.048 & 0.062 & 0.075 & 0.086 & 0.097 \\
    GR00T-N1.6-Bridge         & 57.1 & 0.023 & 0.024 & \textbf{0.026} & 0.029 & 0.031 & 0.038 & 0.051 & 0.064 & 0.076 & 0.091 \\
    CogACT-Base               & 51.3 & 0.031 & 0.031 & \textbf{0.032} & 0.037 & 0.040 & 0.049 & 0.062 & 0.074 & 0.086 & 0.098 \\
    CogACT-Small              & 51.0 & 0.031 & 0.031 & \textbf{0.032} & 0.036 & 0.040 & 0.047 & 0.060 & 0.071 & 0.083 & 0.092 \\
    Pi0-Bridge                & 47.9 & 0.017 & 0.017 & \textbf{0.017} & 0.020 & 0.022 & 0.025 & 0.033 & 0.044 & 0.053 & 0.061 \\
    Qwen2.5-OFT-RT-1-Bridge   & 44.2 & 0.024 & 0.023 & \textbf{0.024} & 0.028 & 0.030 & 0.036 & 0.049 & 0.061 & 0.070 & 0.082 \\
    Qwen3-OFT-RT-1-Bridge     & 42.7 & 0.024 & 0.024 & \textbf{0.025} & 0.029 & 0.031 & 0.038 & 0.051 & 0.064 & 0.075 & 0.086 \\
    SpatialVLA-Bridge         & 42.7 & 0.021 & 0.022 & \textbf{0.022} & 0.024 & 0.026 & 0.033 & 0.044 & 0.054 & 0.063 & 0.073 \\
    GR00T-N1.5-Bridge         & 36.5 & 0.023 & 0.023 & \textbf{0.024} & 0.027 & 0.030 & 0.036 & 0.048 & 0.058 & 0.069 & 0.080 \\
    \midrule
    Spearman $\rho$  & & 0.736 & 0.764 & \textbf{0.802} & 0.788 & 0.767 & 0.767 & 0.774 & 0.746 & 0.774 & 0.725 \\
    $p$    & & 0.003 & 0.002 & \textbf{0.001} & 0.001 & 0.002 & 0.002 & 0.002 & 0.003 & 0.002 & 0.004 \\
    Pearson $r$      & & 0.692 & 0.699 & \textbf{0.717} & 0.696 & 0.689 & 0.683 & 0.647 & 0.646 & 0.629 & 0.610 \\
    $p$    & & 0.006 & 0.006 & \textbf{0.004} & 0.006 & 0.007 & 0.007 & 0.012 & 0.012 & 0.014 & 0.018 \\
    \bottomrule
  \end{tabular}
\par\vspace{4pt}
  \centering
  \scriptsize
  \setlength{\tabcolsep}{3pt}
  \caption{Sensorimotor Alignment (\PA) from vision features at the language-model output versus SimplerEnv-WidowX task success rate (SR) across trajectory horizons $h$.
We score 12 fine-tuned vision-language-action models on DROID ($N=6{,}000$ episodes, $k=10$ neighbors).
Bottom rows report Spearman's $\rho$ and Pearson's $r$ with their one-tailed $p$-values.
Bold marks the horizon of peak Pearson $r$ ($h=15$).}
  \label{tab:cknna-sr-full-img}
  \begin{tabular}{lccccccccccc}
    \toprule
    Model & SR (\%) & $h\!=\!1$ & $h\!=\!3$ & $h\!=\!7$ & $h\!=\!15$ & $h\!=\!25$ & $h\!=\!40$ & $h\!=\!75$ & $h\!=\!110$ & $h\!=\!150$ & $h\!=\!220$ \\
    \midrule
    Qwen2.5-GR00T-Bridge      & 71.4 & 0.034 & 0.034 & 0.035 & \textbf{0.039} & 0.043 & 0.049 & 0.062 & 0.074 & 0.084 & 0.093 \\
    Qwen3-GR00T-RT-1-Bridge   & 65.3 & 0.026 & 0.026 & 0.027 & \textbf{0.031} & 0.034 & 0.041 & 0.054 & 0.067 & 0.080 & 0.092 \\
    Qwen2.5-GR00T-RT-1-Bridge & 63.6 & 0.033 & 0.033 & 0.035 & \textbf{0.038} & 0.041 & 0.047 & 0.059 & 0.070 & 0.079 & 0.088 \\
    CogACT-Large              & 58.3 & 0.030 & 0.030 & 0.032 & \textbf{0.035} & 0.039 & 0.047 & 0.061 & 0.073 & 0.084 & 0.095 \\
    GR00T-N1.6-Bridge         & 57.1 & 0.024 & 0.024 & 0.026 & \textbf{0.029} & 0.032 & 0.039 & 0.051 & 0.064 & 0.077 & 0.091 \\
    CogACT-Base               & 51.3 & 0.031 & 0.030 & 0.032 & \textbf{0.036} & 0.040 & 0.048 & 0.060 & 0.073 & 0.085 & 0.098 \\
    CogACT-Small              & 51.0 & 0.030 & 0.030 & 0.031 & \textbf{0.035} & 0.039 & 0.046 & 0.059 & 0.070 & 0.082 & 0.092 \\
    Pi0-Bridge                & 47.9 & 0.017 & 0.017 & 0.017 & \textbf{0.020} & 0.022 & 0.025 & 0.033 & 0.044 & 0.053 & 0.061 \\
    Qwen2.5-OFT-RT-1-Bridge   & 44.2 & 0.022 & 0.022 & 0.023 & \textbf{0.026} & 0.028 & 0.035 & 0.046 & 0.058 & 0.068 & 0.079 \\
    Qwen3-OFT-RT-1-Bridge     & 42.7 & 0.024 & 0.024 & 0.025 & \textbf{0.028} & 0.030 & 0.037 & 0.050 & 0.063 & 0.074 & 0.087 \\
    SpatialVLA-Bridge         & 42.7 & 0.022 & 0.022 & 0.022 & \textbf{0.025} & 0.026 & 0.033 & 0.045 & 0.055 & 0.064 & 0.075 \\
    GR00T-N1.5-Bridge         & 36.5 & 0.023 & 0.023 & 0.024 & \textbf{0.027} & 0.029 & 0.035 & 0.048 & 0.059 & 0.069 & 0.079 \\
    \midrule
    Spearman $\rho$  & & 0.767 & 0.757 & 0.767 & \textbf{0.746} & 0.746 & 0.739 & 0.725 & 0.711 & 0.683 & 0.704 \\
    $p$    & & 0.002 & 0.002 & 0.002 & \textbf{0.003} & 0.003 & 0.003 & 0.004 & 0.005 & 0.007 & 0.005 \\
    Pearson $r$      & & 0.686 & 0.694 & 0.701 & \textbf{0.703} & 0.690 & 0.647 & 0.606 & 0.618 & 0.601 & 0.554 \\
    $p$    & & 0.007 & 0.006 & 0.006 & \textbf{0.005} & 0.007 & 0.011 & 0.018 & 0.016 & 0.019 & 0.031 \\
    \bottomrule
  \end{tabular}
\par\vspace{4pt}
  \centering
  \scriptsize
  \setlength{\tabcolsep}{3pt}
  \caption{Sensorimotor Alignment (\PA) from text features at the language-model output versus SimplerEnv-WidowX task success rate (SR) across trajectory horizons $h$.
We score 12 fine-tuned vision-language-action models on DROID ($N=6{,}000$ episodes, $k=10$ neighbors).
Bottom rows report Spearman's $\rho$ and Pearson's $r$ with their one-tailed $p$-values.
Bold marks the highest Spearman $\rho$ ($h=150$).\label{tab:cknna-sr-full-txt}}
  \begin{tabular}{lccccccccccc}
    \toprule
    Model & SR (\%) & $h\!=\!1$ & $h\!=\!3$ & $h\!=\!7$ & $h\!=\!15$ & $h\!=\!25$ & $h\!=\!40$ & $h\!=\!75$ & $h\!=\!110$ & $h\!=\!150$ & $h\!=\!220$ \\
    \midrule
    Qwen2.5-GR00T-Bridge      & 71.4 & 0.018 & 0.019 & 0.019 & 0.022 & 0.025 & 0.030 & 0.039 & 0.046 & 0.053 & 0.059 \\
    Qwen3-GR00T-RT-1-Bridge   & 65.3 & 0.015 & 0.015 & 0.016 & 0.018 & 0.020 & 0.024 & 0.031 & 0.040 & 0.048 & 0.057 \\
    Qwen2.5-GR00T-RT-1-Bridge & 63.6 & 0.022 & 0.021 & 0.022 & 0.026 & 0.028 & 0.032 & 0.040 & 0.048 & 0.053 & 0.058 \\
    CogACT-Large              & 58.3 & 0.037 & 0.037 & 0.038 & 0.042 & 0.045 & 0.054 & 0.068 & 0.081 & 0.091 & 0.104 \\
    GR00T-N1.6-Bridge         & 57.1 & 0.008 & 0.008 & 0.008 & 0.010 & 0.010 & 0.014 & 0.020 & 0.024 & 0.029 & 0.033 \\
    CogACT-Base               & 51.3 & 0.035 & 0.035 & 0.037 & 0.041 & 0.044 & 0.053 & 0.068 & 0.080 & 0.090 & 0.104 \\
    CogACT-Small              & 51.0 & 0.030 & 0.029 & 0.030 & 0.034 & 0.038 & 0.044 & 0.057 & 0.064 & 0.070 & 0.080 \\
    Pi0-Bridge                & 47.9 & 0.014 & 0.015 & 0.016 & 0.019 & 0.021 & 0.024 & 0.033 & 0.043 & 0.052 & 0.061 \\
    Qwen2.5-OFT-RT-1-Bridge   & 44.2 & 0.019 & 0.018 & 0.018 & 0.022 & 0.024 & 0.028 & 0.036 & 0.044 & 0.051 & 0.060 \\
    Qwen3-OFT-RT-1-Bridge     & 42.7 & 0.013 & 0.013 & 0.014 & 0.015 & 0.016 & 0.020 & 0.026 & 0.034 & 0.040 & 0.045 \\
    SpatialVLA-Bridge         & 42.7 & 0.014 & 0.014 & 0.014 & 0.016 & 0.017 & 0.022 & 0.030 & 0.036 & 0.043 & 0.049 \\
    GR00T-N1.5-Bridge         & 36.5 & 0.010 & 0.010 & 0.010 & 0.011 & 0.011 & 0.014 & 0.019 & 0.024 & 0.029 & 0.033 \\
    \midrule
    Spearman $\rho$  & & 0.438 & 0.452 & 0.452 & 0.452 & 0.452 & 0.452 & 0.501 & 0.452 & \textbf{0.508} & 0.291 \\
    $p$    & & 0.077 & 0.070 & 0.070 & 0.070 & 0.070 & 0.070 & 0.049 & 0.070 & \textbf{0.046} & 0.180 \\
    Pearson $r$      & & 0.227 & 0.238 & 0.245 & 0.265 & 0.274 & 0.259 & 0.254 & 0.265 & 0.255 & 0.233 \\
    $p$    & & 0.239 & 0.229 & 0.222 & 0.203 & 0.194 & 0.208 & 0.213 & 0.202 & 0.212 & 0.233 \\
    \bottomrule
  \end{tabular}
\end{table}

\FloatBarrier
\subsection{Implementation Choices}
\label{app:implementation-sensitivity}
\label{app:k-sweep}

We vary one choice at a time, keeping the main-paper defaults otherwise: the 6,000-episode DROID subset, $h\!=\!7$, $k\!=\!10$, the joint vision-text features at the language-model output, and the 12 finetuned VLAs. \cref{tab:sensitivity} reports the Pearson $r$ between \mbox{\PA} and SimplerEnv success rate for each setting. All settings of $k$, normalization, component weights, rotation distance, and pooling keep the correlation positive and significant, with $r$ between $+0.634$ and $+0.752$.

\begin{table}[!htbp]
  \centering
  \footnotesize
  \caption{\textbf{Sensitivity of the Sensorimotor Alignment (\PA)--success-rate correlation to implementation choices.}
Pearson's $r$ and one-tailed $p$ compare \PA with SimplerEnv-WidowX task success rate across 12 fine-tuned vision-language-action models.
Each block varies one choice; $^\ast$ marks the default.\label{tab:sensitivity}}
  \setlength{\tabcolsep}{4pt}
  \begin{minipage}[t]{0.48\textwidth}\vspace{0pt}
    \centering
    \begin{tabular}{lcc}
      \toprule
      Setting & $r$ & $p$ \\
      \midrule
      \multicolumn{3}{l}{\textit{Number of neighbors $k$}} \\
      $k\!=\!2$            & $+0.679$ & $0.0076$ \\
      $k\!=\!5$            & $+0.667$ & $0.0089$ \\
      $k\!=\!10^\ast$      & $+0.717$ & $0.0043$ \\
      $k\!=\!15$           & $+0.665$ & $0.0091$ \\
      $k\!=\!20$           & $+0.644$ & $0.0119$ \\
      \midrule
      \multicolumn{3}{l}{\textit{Trajectory normalization}} \\
      Absolute pose$^\ast$ & $+0.717$ & $0.0043$ \\
      Anchor-relative      & $+0.752$ & $0.0024$ \\
      \midrule
      \multicolumn{3}{l}{\textit{Rotation distance}} \\
      $\SO(3)$ geodesic$^\ast$   & $+0.717$ & $0.0043$ \\
      Quaternion squared chordal & $+0.686$ & $0.0069$ \\
      XYZ Euler $\ell_2$            & $+0.660$ & $0.0097$ \\
      \midrule
      \multicolumn{3}{l}{\textit{Token pooling}} \\
      Masked mean$^\ast$   & $+0.717$ & $0.0043$ \\
      Masked max           & $+0.653$ & $0.0107$ \\
      Masked min           & $+0.730$ & $0.0035$ \\
      \bottomrule
    \end{tabular}
  \end{minipage}\hfill
  \begin{minipage}[t]{0.48\textwidth}\vspace{0pt}
    \centering
    \begin{tabular}{lcc}
      \toprule
      Setting & $r$ & $p$ \\
      \midrule
      \multicolumn{3}{l}{\textit{Weights (position, rotation, gripper)}} \\
      $(1, 1, 1)^\ast$   & $+0.717$ & $0.0043$ \\
      $(0.5, 1, 1)$      & $+0.698$ & $0.0058$ \\
      $(2, 1, 1)$        & $+0.708$ & $0.0050$ \\
      $(1, 0.5, 1)$      & $+0.694$ & $0.0062$ \\
      $(1, 2, 1)$        & $+0.703$ & $0.0054$ \\
      $(1, 1, 0.5)$      & $+0.702$ & $0.0055$ \\
      $(1, 1, 2)$        & $+0.699$ & $0.0057$ \\
      $(0, 1, 1)$        & $+0.745$ & $0.0027$ \\
      $(1, 0, 1)$        & $+0.658$ & $0.0100$ \\
      $(1, 1, 0)$        & $+0.634$ & $0.0135$ \\
      \bottomrule
    \end{tabular}
  \end{minipage}
\end{table}

\textbf{Number of neighbors $k$.}
We use $k\!=\!10$ in the main paper, following \citet{huh2024platonic}. Across $k\!\in\!\{2,5,10,15,20\}$, $r$ ranges from $+0.644$ to $+0.717$, with a mild peak at the default $k\!=\!10$.

\textbf{Trajectory normalization.}
The anchor-relative setting replaces each position with its offset from the starting frame and keeps rotation and gripper state unchanged. It gives a slightly higher $r$ than absolute poses, possibly because removing the workspace offset emphasizes the shape of the motion. We keep absolute poses for simplicity.

\textbf{Component weights.}
The DTW step cost in Appendix~\ref{app:base-frame} combines position, rotation, and gripper-state distances with weights $w_{\mathrm{pos}}$, $w_{\mathrm{rot}}$, and $w_{\mathrm{grip}}$, all equal to one by default. Halving, doubling, or removing one weight keeps $r$ between $+0.634$ and $+0.745$. Removing rotation or gripper lowers $r$ most, suggesting both carry information beyond position.

\textbf{Rotation distance and token pooling.}
The $\SO(3)$ geodesic gives the highest $r$ of the three rotation distances. The quaternion alternative uses the scaled squared chordal quantity \mbox{$1-|\mathbf{q}^{\top}\mathbf{q}^{\prime}|$}; the Euler alternative uses the $\ell_2$ distance between recorded angle triples, without wrapping angular differences. Each alternative is rescaled to match the mean geodesic distance on 40,000 sampled state pairs. For pooling the LLM-output tokens, masked min gives a slightly higher $r$ than the default masked mean, and masked max gives the lowest.

\FloatBarrier
\subsection{DTW Alignment}
\label{app:dtw-ablation}

\mbox{\PA} aligns trajectory pairs with DTW before measuring their distance, so the same motion at different speeds counts as similar. To test whether the correlation depends on this warping, we compare DTW with the lockstep distance $d_{\mathrm{lock}}(i,j)=\sum_{u=1}^{h}c_{uu}$, which matches frames by time index using the local cost in Appendix~\ref{app:base-frame}. The lockstep sum weights every matched state once, whereas symmetric2 DTW weights diagonal transitions twice. \cref{tab:dtw-ablation} reports the comparison. DTW raises \mbox{\PA} for all 12 VLAs from $h\!=\!7$ onward, and the gain grows with the horizon, from $1.2\%$ at $h\!=\!7$ to $34.1\%$ at $h\!=\!220$, since longer trajectories contain more variation in speed. Because DTW raises the \mbox{\PA} of all models by a similar amount, $(34\pm4)\%$ (mean $\pm$ standard deviation across models) at $h\!=\!220$, the correlation with task success barely changes. Across the 10 horizons, $r$ ranges from \mbox{${+}0.617$} to \mbox{${+}0.713$} without DTW and from \mbox{${+}0.610$} to \mbox{${+}0.717$} with DTW, a relative difference between \mbox{$-3.2\%$} and \mbox{$+0.6\%$}. The \mbox{\PA}--success-rate correlation is therefore not an artifact of warping.

\begin{table}[!htbp]
  \centering
  \caption{\textbf{Comparing Sensorimotor Alignment (\PA) with and without Dynamic Time Warping (DTW).}
DTW raises absolute \PA more at longer trajectory horizons $h$, while its Pearson correlation $r$ with SimplerEnv-WidowX task success rate (SR) stays stable.
Each relative change is $100(v_{\mathrm{DTW}}/v_{\mathrm{no\,DTW}}-1)\%$ for the quantity $v$ in that block.\label{tab:dtw-ablation}}
  \small
  \setlength{\tabcolsep}{6pt}
  \begin{tabular}{c ccc ccc}
    \toprule
    & \multicolumn{3}{c}{\PA\ (mean over 12 VLAs)} & \multicolumn{3}{c}{\PA--SR Pearson $r$} \\
    \cmidrule(lr){2-4}\cmidrule(lr){5-7}
    Horizon $h$ & w/ DTW & w/o DTW & rel.\ change & w/ DTW & w/o DTW & rel.\ change \\
    \midrule
    $1$   & $0.0268$ & $0.0268$ & $+0.0\%$  & $+0.692$ & $+0.692$ & $+0.0\%$ \\
    $3$   & $0.0268$ & $0.0269$ & $-0.6\%$  & $+0.699$ & $+0.700$ & $-0.1\%$ \\
    $7$   & $0.0280$ & $0.0277$ & $+1.2\%$  & $+0.717$ & $+0.713$ & $+0.6\%$ \\
    $15$  & $0.0314$ & $0.0302$ & $+4.1\%$  & $+0.696$ & $+0.710$ & $-2.0\%$ \\
    $25$  & $0.0342$ & $0.0317$ & $+7.7\%$  & $+0.689$ & $+0.711$ & $-3.2\%$ \\
    $40$  & $0.0410$ & $0.0365$ & $+12.4\%$ & $+0.683$ & $+0.680$ & $+0.4\%$ \\
    $75$  & $0.0537$ & $0.0442$ & $+21.7\%$ & $+0.647$ & $+0.663$ & $-2.5\%$ \\
    $110$ & $0.0653$ & $0.0533$ & $+22.4\%$ & $+0.646$ & $+0.661$ & $-2.2\%$ \\
    $150$ & $0.0758$ & $0.0602$ & $+25.9\%$ & $+0.629$ & $+0.633$ & $-0.7\%$ \\
    $220$ & $0.0870$ & $0.0649$ & $+34.1\%$ & $+0.610$ & $+0.617$ & $-1.1\%$ \\
    \bottomrule
  \end{tabular}
\end{table}

\FloatBarrier
\subsection{Sensitivity to Embodiment Mismatch}
\label{app:embodiment-mismatch}

\begin{figure}[!htbp]

  \captionsetup{justification=raggedright}
  \centering

  \begin{minipage}{0.55\textwidth}
    \includegraphics[width=0.99\linewidth]{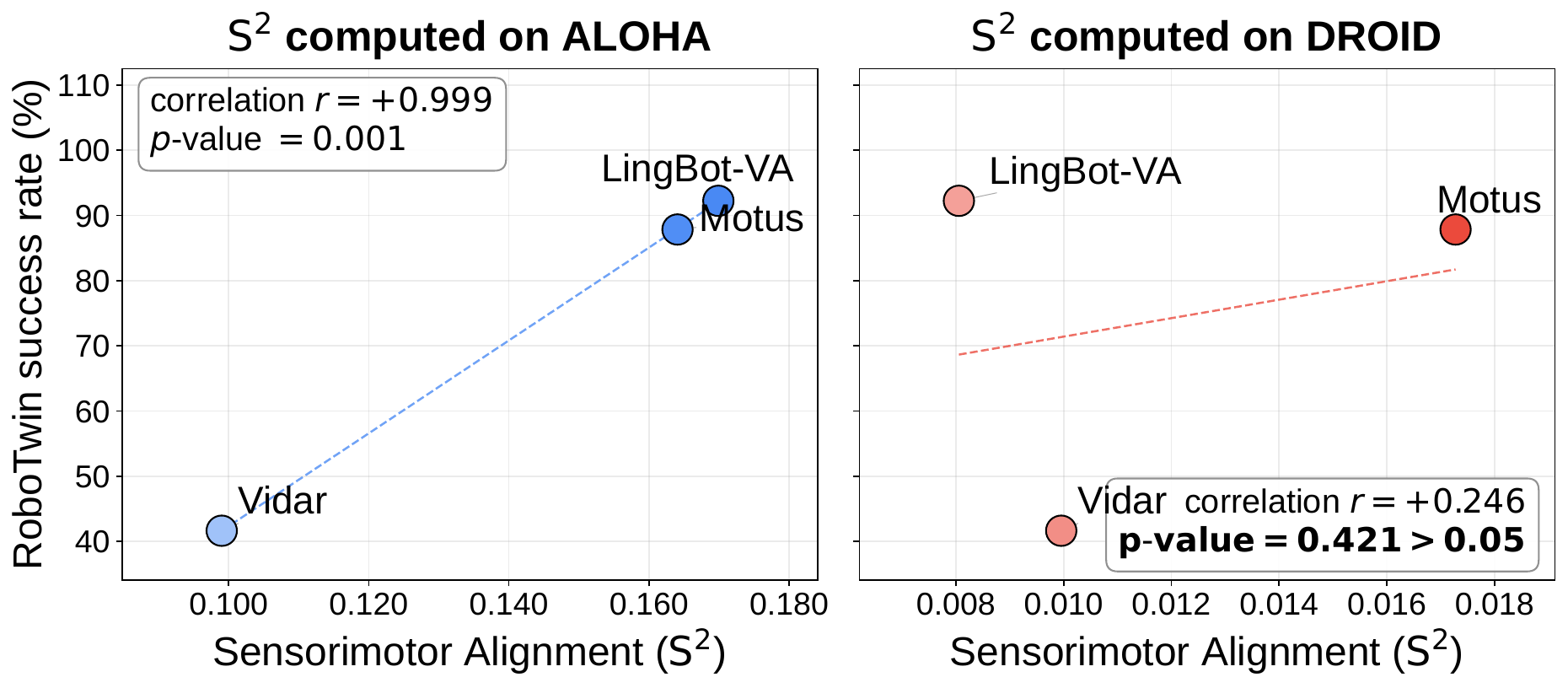}
  \end{minipage}\hfill
  \begin{minipage}{0.4\textwidth}

    \caption{\textbf{Sensorimotor Alignment (\PA) is sensitive to a mismatch in arm configuration.}
For 3 world action models (WAMs) evaluated on bimanual RoboTwin, the matched bimanual ALOHA reference (left) recovers the success-rate ordering, while the single-arm DROID reference (right) does not.\label{fig:va-sr-aloha-vs-droid}}
  \end{minipage}

\end{figure}

For the negative control in \cref{sec:finding-robustness}, we score 3 bimanual WAMs evaluated on RoboTwin with two reference datasets that differ in arm configuration, as shown in \cref{fig:va-sr-aloha-vs-droid}. With the matched bimanual ALOHA reference~\citep{zhao2023learning}, \mbox{\PA} recovers the success-rate order of all 3 models, at \mbox{$r\!=\!{+}0.999$}. With the single-arm DROID reference, the correlation drops to \mbox{$r\!=\!{+}0.246$} with \mbox{$p\!=\!0.421$}. \mbox{\PA} therefore responds to the mismatch in arm configuration rather than reporting alignment indiscriminately. This sensitivity complements the stability across single-arm references in \cref{sec:finding-robustness}, where DROID and BridgeDataV2 differ in robot, scenes, and tasks yet give nearly the same correlation for the 12 VLAs. A reference dataset therefore need not come from the exact target robot, but it should match the target's arm configuration.

\FloatBarrier
\section{Details of the Representation-Level Explanations}
\label{app:diagnostics}

This section gives the full data behind the representation-level explanations in \cref{sec:pa-reveals}.

\subsection{\texorpdfstring{\PA}{S2} Across Feature Positions}
\label{app:full-cknna-per-model}

\cref{sec:finding-what-carries} compares \mbox{\PA} from vision, text, and joint features at the LLM output. Here we extend the comparison to every feature position, from the vision-encoder output (ViT-out) through the language-model input (LLM-in) to the language-model output (LLM-out).

\textbf{Correlation with task success.}
\cref{tab:feature-position-r} reports the correlation between \mbox{\PA} and SimplerEnv success rate for the 12 fine-tuned VLAs at each feature position on DROID ($N\!=\!6{,}000$, $h\!=\!7$); \cref{sec:finding-what-carries} reports $h\!=\!3$, and Appendix~\ref{app:cknna-sr-full} gives every horizon. The correlation holds at every visual feature position and grows with depth, from \mbox{$r\!=\!{+}0.567$} at the vision-encoder output to \mbox{$r\!=\!{+}0.717$} at the LLM output. Text-only features show no significant correlation at either position. This agrees with reports that VLAs follow visual cues despite conflicting instructions~\citep{libero-cf,igar}.

\begin{table}[!htbp]
  \centering
  \small
  \caption{\textbf{Correlation between Sensorimotor Alignment (\PA) and task success by feature position.}
Pearson's $r$ and one-tailed $p$ compare \PA with SimplerEnv-WidowX task success rate across 12 fine-tuned vision-language-action models on DROID ($N=6{,}000$ episodes, trajectory horizon $h=7$).
ViT-out, LLM-in, and LLM-out denote the vision-encoder output, language-model input, and language-model output; $^\ast$ marks the default feature.\label{tab:feature-position-r}}
  \begin{tabular}{lcc}
    \toprule
    Feature position & Pearson's $r$ & One-tailed $p$ \\
    \midrule
    ViT-out                        & $+0.567$ & $0.0274$ \\
    LLM-in vision-text             & $+0.580$ & $0.0239$ \\
    LLM-out vision-only            & $+0.701$ & $0.0056$ \\
    LLM-out vision-text$^\ast$     & $+0.717$ & $0.0043$ \\
    LLM-in text-only               & $-0.325$ & $0.8486$ \\
    LLM-out text-only              & $+0.245$ & $0.2215$ \\
    \bottomrule
  \end{tabular}
\end{table}

\textbf{\PA by model group.}
\cref{tab:raw-vs-finetuned-cknna-mean} reports the mean \mbox{\PA} at each feature position for 3 model groups on DROID ($N\!=\!6{,}000$, $h\!=\!7$): 3 standalone vision encoders, the 8 raw VLMs, and the 12 fine-tuned VLAs. \cref{tab:full-cknna-per-model} gives the per-model values. For the vision, text, and joint vision-text features of \cref{sec:finding-what-carries}, we average-pool the selected output tokens and apply the feature normalization in Appendix~\ref{app:cknna-estimator}. The VLA group has higher mean \mbox{\PA} than the raw-VLM group at every compared position. The relative difference is larger at the LLM output, from $+29.9\%$ to $+45.0\%$, than at the vision-encoder output and LLM input, from $+12.7\%$ to $+18.1\%$.

\begin{table}[!htbp]
    \centering
    \caption{\textbf{Mean Sensorimotor Alignment (\PA) by feature position} on DROID ($N=6{,}000$ episodes, trajectory horizon $h=7$).
ViT-out, LLM-in, and LLM-out denote the vision-encoder output, language-model input, and language-model output; standalone vision encoders have only ViT-out features.
The last column is $100(\overline{\mathrm{S}^2}_{\mathrm{VLA}}/\overline{\mathrm{S}^2}_{\mathrm{VLM}}-1)\%$, where each bar denotes the model-group mean.\label{tab:raw-vs-finetuned-cknna-mean}}
    \small
    \begin{tabular}{lcccr}
        \toprule
        \textbf{Feature} & \textbf{Vision encoder} & \textbf{Raw VLM} & \textbf{Finetuned VLA} & \textbf{Relative change} \\
                         & \textbf{(mean, $n\!=\!3$)} & \textbf{(mean, $n\!=\!8$)} & \textbf{(mean, $n\!=\!12$)} & \\
        \midrule
        ViT-out                          & $0.0246$ & $0.0247$ & $0.0278$ & $+12.7\%$ \\
        LLM-in vision-only                         & \textendash & $0.0244$ & $0.0275$ & $+12.7\%$ \\
        LLM-in vision-text                       & \textendash & $0.0233$ & $0.0275$ & $+18.1\%$ \\
        LLM-out vision-only                        & \textendash & $0.0212$ & $0.0275$ & $+29.9\%$ \\
        LLM-out text-only                        & \textendash & $0.0140$ & $0.0202$ & $+45.0\%$ \\
        LLM-out vision-text                      & \textendash & $0.0204$ & $0.0280$ & $+37.4\%$ \\
        \bottomrule
    \end{tabular}
\end{table}

\begin{table}[!htbp]
    \centering
    \caption{\textbf{Per-model Sensorimotor Alignment (\PA) by feature position} on DROID ($N=6{,}000$ episodes, trajectory horizon $h=7$).
ViT-out, LLM-in, and LLM-out denote the vision-encoder output, language-model input, and language-model output.
Vision-only, text-only, and vision-text select image tokens, instruction tokens, or both from the same model input.
Group means are in \cref{tab:raw-vs-finetuned-cknna-mean}.\label{tab:full-cknna-per-model}}
    \scriptsize
    \setlength{\tabcolsep}{3pt}
    \begin{tabular}{l ccccccc}
        \toprule
        \textbf{Model} & \textbf{ViT-out} & \makecell{\textbf{LLM-in}\\\textbf{vision-only}} & \makecell{\textbf{LLM-in}\\\textbf{text-only}} & \makecell{\textbf{LLM-in}\\\textbf{vision-text}} & \makecell{\textbf{LLM-out}\\\textbf{vision-only}} & \makecell{\textbf{LLM-out}\\\textbf{text-only}} & \makecell{\textbf{LLM-out}\\\textbf{vision-text}} \\
        \midrule
        \multicolumn{8}{l}{\textit{Pretrained vision encoders, no LLM attached ($n\!=\!3$)}} \\
        \midrule
        CLIP-ViT-L/14~\citep{radford2021learning} & $0.0240$ & \textendash & \textendash & \textendash & \textendash & \textendash & \textendash \\
        SigLIP-SO400M~\citep{zhai2023sigmoid} & $0.0247$ & \textendash & \textendash & \textendash & \textendash & \textendash & \textendash \\
        SigLIP-2 Large~\citep{tschannen2025siglip2} & $0.0250$ & \textendash & \textendash & \textendash & \textendash & \textendash & \textendash \\
        \midrule
        \multicolumn{8}{l}{\textit{Raw VLMs from VLM4VLA ($n\!=\!8$)}} \\
        \midrule
        KosMos-2                 & $0.0257$ & $0.0238$ & $0.0093$ & $0.0239$ & $0.0214$ & $0.0151$ & $0.0216$ \\
        PaliGemma 1              & $0.0249$ & $0.0251$ & $0.0082$ & $0.0193$ & $0.0193$ & $0.0160$ & $0.0191$ \\
        PaliGemma 2              & $0.0250$ & $0.0251$ & $0.0087$ & $0.0222$ & $0.0196$ & $0.0142$ & $0.0192$ \\
        Qwen2.5-VL-3B            & $0.0241$ & $0.0239$ & $0.0079$ & $0.0239$ & $0.0185$ & $0.0139$ & $0.0188$ \\
        Qwen2.5-VL-7B            & $0.0231$ & $0.0234$ & $0.0083$ & $0.0234$ & $0.0188$ & $0.0127$ & $0.0190$ \\
        Qwen3-VL-2B              & $0.0242$ & $0.0237$ & $0.0089$ & $0.0237$ & $0.0236$ & $0.0127$ & $0.0217$ \\
        Qwen3-VL-4B              & $0.0253$ & $0.0254$ & $0.0089$ & $0.0254$ & $0.0239$ & $0.0135$ & $0.0232$ \\
        Qwen3-VL-8B              & $0.0250$ & $0.0245$ & $0.0088$ & $0.0245$ & $0.0240$ & $0.0136$ & $0.0207$ \\
        \midrule
        \multicolumn{8}{l}{\textit{Finetuned VLAs ($n\!=\!12$)}} \\
        \midrule
        Qwen2.5-GR00T-Bridge           & $0.0344$ & $0.0335$ & $0.0078$ & $0.0335$ & $0.0354$ & $0.0192$ & $0.0363$ \\
        Qwen2.5-GR00T-RT-1-Bridge      & $0.0355$ & $0.0345$ & $0.0077$ & $0.0345$ & $0.0350$ & $0.0224$ & $0.0353$ \\
        Qwen2.5-OFT-RT-1-Bridge        & $0.0265$ & $0.0265$ & $0.0082$ & $0.0265$ & $0.0234$ & $0.0185$ & $0.0242$ \\
        Qwen3-GR00T-RT-1-Bridge        & $0.0264$ & $0.0266$ & $0.0085$ & $0.0266$ & $0.0274$ & $0.0158$ & $0.0292$ \\
        Qwen3-OFT-RT-1-Bridge          & $0.0263$ & $0.0257$ & $0.0086$ & $0.0257$ & $0.0248$ & $0.0136$ & $0.0254$ \\
        CogACT Small                   & $0.0301$ & $0.0303$ & $0.0097$ & $0.0303$ & $0.0313$ & $0.0300$ & $0.0319$ \\
        CogACT Base                    & $0.0301$ & $0.0298$ & $0.0096$ & $0.0298$ & $0.0319$ & $0.0368$ & $0.0324$ \\
        CogACT Large                   & $0.0298$ & $0.0298$ & $0.0097$ & $0.0298$ & $0.0317$ & $0.0379$ & $0.0328$ \\
        GR00T-N1.5-Bridge              & $0.0252$ & $0.0244$ & $0.0090$ & $0.0243$ & $0.0238$ & $0.0099$ & $0.0240$ \\
        GR00T-N1.6-Bridge              & $0.0251$ & $0.0245$ & $0.0090$ & $0.0245$ & $0.0258$ & $0.0082$ & $0.0255$ \\
        Pi0-Bridge                     & $0.0176$ & $0.0178$ & $0.0090$ & $0.0180$ & $0.0174$ & $0.0161$ & $0.0174$ \\
        SpatialVLA-Bridge              & $0.0266$ & $0.0263$ & $0.0087$ & $0.0265$ & $0.0219$ & $0.0145$ & $0.0217$ \\
        \bottomrule
    \end{tabular}
\end{table}

\FloatBarrier
\subsection{Frozen Versus Unfrozen Vision Encoder}
\label{app:freeze-ablation-horizon-sweep}

Following VLM4VLA~\citep{vlm4vla}, both VLAs in \cref{sec:pa-reveals-frozen} attach an FCDecoder action head to Qwen2.5-VL-3B~\citep{bai2025qwen25vltechnicalreport} and are fine-tuned on BridgeDataV2. Training differs only in whether the vision encoder is frozen. We compare each configuration's checkpoint with the best observed success rate: step 10,000 for the unfrozen model and step 30,000 for the frozen model. \cref{tab:freeze-ablation-horizon-sweep} compares all 10 horizons using mean-pooled LLM-output vision-text features, as in the main experiments, rather than the action-token state read by the action head. The frozen model has lower \mbox{\PA} at every horizon, with relative reductions from \mbox{$38.1\%$} at \mbox{$h\!=\!1$} to \mbox{$49.4\%$} at \mbox{$h\!=\!220$}.

\begin{table}[!htbp]
    \centering
    \caption{\textbf{Sensorimotor Alignment (\PA) with a frozen and an unfrozen vision encoder across trajectory horizons $h$.}
Qwen2.5-VL-3B vision-language-action models trained with the same recipe, evaluated using joint vision-text features at the language-model output on DROID ($N=6{,}000$ episodes).
Relative change is $100(\mathrm{S}^2_{\mathrm{frozen}}/\mathrm{S}^2_{\mathrm{unfrozen}}-1)\%$.\label{tab:freeze-ablation-horizon-sweep}}
    \small
    \begin{tabular}{rccr}
        \toprule
        \textbf{Horizon $h$} & \textbf{Unfrozen \PA} & \textbf{Frozen \PA} & \textbf{Relative change} \\
        \midrule
        $1$   & $0.0231$ & $0.0143$ & $-38.1\%$ \\
        $3$   & $0.0232$ & $0.0141$ & $-39.0\%$ \\
        $7$   & $0.0245$ & $0.0144$ & $-41.1\%$ \\
        $15$  & $0.0268$ & $0.0158$ & $-41.2\%$ \\
        $25$  & $0.0295$ & $0.0166$ & $-43.8\%$ \\
        $40$  & $0.0352$ & $0.0191$ & $-45.7\%$ \\
        $75$  & $0.0481$ & $0.0258$ & $-46.4\%$ \\
        $110$ & $0.0614$ & $0.0326$ & $-46.9\%$ \\
        $150$ & $0.0733$ & $0.0390$ & $-46.8\%$ \\
        $220$ & $0.0861$ & $0.0435$ & $-49.4\%$ \\
        \bottomrule
    \end{tabular}
\end{table}

\FloatBarrier
\Needspace{610pt}
\subsection{WAM Features Along the Denoising Schedule}
\label{app:layerwise-vs-denoising-step}

\cref{tab:layerwise-vs-denoising-step} reports \mbox{\PA} for 3 WAMs at 7 feature positions and 5 points of each model's denoising schedule. \cref{fig:pa-vs-denoising-pct} plots the final-layer-norm row, and \cref{fig:pa-layerwise} plots each row's schedule average. Block~14 has the highest \mbox{\PA} at 13 of the 15 model and schedule points. Columns give Motus steps 1, 3, 5, 8, and 10 of 10; LingBot-VA steps 0, 6, 13, 19, and 25 of 25; and Vidar steps 0, 5, 10, 15, and 19 of 20; the last Motus and Vidar features precede the final update.

\begin{table}[H]
\centering
\caption{\textbf{Layer-wise Sensorimotor Alignment (\PA) along the denoising schedule for 3 world action models (WAMs).}
Columns label 5 selected positions in each model's native schedule, not matched noise levels.
Motus and LingBot-VA use 500 episodes over 50 RoboTwin tasks; Vidar uses 480 episodes over 48 tasks.
Rows give 6 video-backbone positions and the text embedding.
Block~14 (bold) has the highest \PA at 13 of the 15 model and schedule positions.
The random baseline is the mean $\pm$ standard deviation (divisor 10) over 10 draws of independent standard-normal features with the same shape.\label{tab:layerwise-vs-denoising-step}}
\small
\setlength{\tabcolsep}{5pt}
\begin{tabular}{@{}l ccccc@{}}
  \toprule
  Feature
    & 0\,\%
    & 25\,\%
    & 50\,\%
    & 75\,\%
    & 100\,\% \\
    & {\scriptsize earliest} & & & & {\scriptsize latest} \\
  \midrule
  \multicolumn{6}{l}{\emph{Motus} \hfill random baseline $= 0.012 \pm 0.002$} \\
  Block 0                    & 0.044 & 0.203 & 0.307 & 0.321 & 0.324 \\
  Block 7                    & 0.342 & 0.388 & 0.399 & 0.390 & 0.378 \\
  \textbf{Block 14}          & \textbf{0.479} & \textbf{0.492} & \textbf{0.482}
                             & \textbf{0.479} & \textbf{0.471} \\
  Block 21                   & 0.454 & 0.448 & 0.447 & 0.436 & 0.424 \\
  Block 29                   & 0.420 & 0.424 & 0.419 & 0.421 & 0.405 \\
  Final layer norm           & 0.425 & 0.421 & 0.421 & 0.418 & 0.402 \\
  Text embedding             & 0.325 & 0.325 & 0.325 & 0.325 & 0.325 \\
  \midrule
  \multicolumn{6}{l}{\emph{LingBot-VA} \hfill random baseline $= 0.012 \pm 0.001$} \\
  Block 0                    & 0.142 & 0.160 & 0.238 & 0.326 & 0.384 \\
  Block 7                    & 0.404 & 0.399 & 0.431 & 0.449 & 0.455 \\
  \textbf{Block 14}          & \textbf{0.498} & \textbf{0.504} & \textbf{0.514}
                             & \textbf{0.514} & \textbf{0.513} \\
  Block 21                   & 0.422 & 0.435 & 0.428 & 0.445 & 0.449 \\
  Block 29                   & 0.423 & 0.440 & 0.441 & 0.447 & 0.448 \\
  Final layer norm           & 0.425 & 0.429 & 0.431 & 0.439 & 0.451 \\
  Text embedding             & 0.355 & 0.355 & 0.355 & 0.355 & 0.355 \\
  \midrule
  \multicolumn{6}{l}{\emph{Vidar} \hfill random baseline $= 0.013 \pm 0.004$} \\
  Block 0                    & 0.483 & 0.477 & 0.459 & 0.429 & 0.422 \\
  Block 7                    & 0.467 & 0.484 & 0.485 & 0.479 & 0.469 \\
  \textbf{Block 14}          & \textbf{0.504} & \textbf{0.555} & \textbf{0.531}
                             & \textbf{0.518} & \textbf{0.473} \\
  Block 21                   & 0.508 & 0.510 & 0.500 & 0.488 & 0.475 \\
  Block 29                   & 0.487 & 0.490 & 0.489 & 0.481 & 0.474 \\
  Final layer norm           & 0.487 & 0.491 & 0.491 & 0.481 & 0.476 \\
  Text embedding             & 0.467 & 0.467 & 0.467 & 0.467 & 0.467 \\
  \bottomrule
\end{tabular}
\end{table}

\textbf{Behavioral validation.}
To check that \mbox{\PA} along the denoising schedule correlates with task success, we vary the number of video denoising steps in LingBot-VA~\citep{li2026causal}. We measure \mbox{\PA} and closed-loop success rate on the RoboTwin \texttt{turn\_switch} task over 100 episodes. Reducing the steps from the default 25 to 5 lowers the success rate from $56\%$ to $48\%$ and \mbox{\PA} from $0.288$ to $0.241$, so \mbox{\PA} and task success change together.

\end{document}